\documentclass{siamart1116}

\usepackage{lipsum}
\usepackage{amsfonts}
\usepackage{graphicx}
\usepackage{epstopdf}
\usepackage{algorithmic}
\usepackage[nolist,nohyperlinks]{acronym}
\usepackage{graphicx}
\usepackage{tikz}
\usepackage{grffile}
\usepackage{xr}
\usepackage{subcaption}
\usepackage{breqn}
\usepackage{amsopn}

\ifpdf
  \DeclareGraphicsExtensions{.eps,.pdf,.png,.jpg}
\else
  \DeclareGraphicsExtensions{.eps}
\fi

\numberwithin{theorem}{section}

\newcommand{\TheTitle}{Intrinsic Isometric Manifold Learning with Application to Localization} 
\newcommand{\TheAuthors}{A. Schwartz and R. Talmon}

\headers{Intrinsic Isometric Manifold Learning}{\TheAuthors}

\title{{\TheTitle}\thanks{Submitted to the editors DATE.}}

\author{
  Ariel Schwartz\thanks{Viterbi Faculty of Electrical Engineering, Technion -- Israel Institute of Technology, Israel
    (\email{ariels@technion.campus.ac.il}}
  \and
  Ronen Talmon\thanks{Viterbi Faculty of Electrical Engineering, Technion -- Israel Institute of Technology, Israel  (\email{ronen@ee.technion.ac.il}}
}

\ifpdf
\hypersetup{
	pdftitle={\TheTitle},
	pdfauthor={\TheAuthors}
}
\fi

\begin{document}
	\begin{acronym}
		\acro{ADAM}{ADAptive Moment estimation}
		\acro{ANN}{Artificial Neural Network}
		\acro{AWGN}{Additive White Gaussian Noise}
		\acro{EDM}{Euclidean Distance Matrix}
		\acro{EDMCP}{Euclidean Distance Matrix Completion Problem}
		\acro{GMM}{Gaussian Mixture Model}
		\acro{LLE}{Locally Linear Embedding}
		\acro{LS-MDS}{Least Squares Multi Dimensional Scaling}
		\acro{MDS}{Multi Dimensional Scaling}
		\acro{MVU}{Maximum Variance Unfolding}
		\acro{PCA}{Principal Component Analysis}
		\acro{RSS}{Received Signal Strength}
		\acro{SDE}{Stochastic Differential Equation}
		\acro{SDP}{SemiDefinite Programming}
		\acro{SMACOF}{Scaling by MAjorizing a COmplicated Function}
		\acro{SNL} {Sensor network localization}
		\acro{SVD}{Singular Value Decomposition}
	\end{acronym} 
	
	\maketitle
	
	\begin{abstract}
	Data living on manifolds commonly appear in many applications. Often this results from an inherently latent low-dimensional system being observed through higher dimensional measurements. We show that under certain conditions, it is possible to construct an intrinsic and isometric data representation, which respects an underlying latent intrinsic geometry. Namely, we view the observed data only as a proxy and learn the structure of a latent unobserved intrinsic manifold, whereas common practice is to learn the manifold of the observed data. For this purpose, we build a new metric and propose a method for its robust estimation by assuming mild statistical priors and by using artificial neural networks as a mechanism for metric regularization and parametrization. We show successful application to unsupervised indoor localization in ad-hoc sensor networks. Specifically, we show that our proposed method facilitates accurate localization of a moving agent from imaging data it collects. Importantly, our method is applied in the same way to two different imaging modalities, thereby demonstrating its intrinsic and modality-invariant capabilities.
	\end{abstract}
	
	\begin{keywords}
		manifold learning, intrinsic, isometric, metric estimation, inverse problem, sensor invariance,  positioning
	\end{keywords}
	
	\begin{AMS}
		57R40, 57M50, 65J22, 57R50  
	\end{AMS}

	\externaldocument{paper.tex}

\section{Introduction}
\label{sec:introduction}

Making measurements is an integral part of everyday life, and particularly, exploration and learning. However, we are usually not interested in the values of the acquired measurements themselves, but rather in understanding the latent system which drives these measurements and might not be directly observable. For example when considering a radar system, we are not interested in the pattern of the electromagnetic wave received and measured by the antenna, but rather in the location, size and velocity of the object captured by the wave reflection pattern. This example highlights the difference between \textit{``observed''} properties, which can be directly measured, and \textit{``intrinsic''} properties corresponding to the latent driving system. The importance of this distinction becomes central when the same latent system is observed and represented in multiple forms; the different measurements and observation modalities may have different observed properties, yet they share certain common intrinsic properties and structures, since they are all generated by the same driving system.
	

Different observation modalities are not all equally suitable for examining the underlying driving system. It is often the case that systems, which are governed by a small number of variables (and are hence inherently low-dimensional in nature), are observed via redundant complex high-dimensional measurements, obscuring the true low-dimensional and much simpler nature of the latent system of interest. As a result, when trying to tackle signal processing and machine learning tasks, a considerable amount of effort is put initially on choosing or building an appropriate representation of the observed data. 

	
Much research in the fields of machine learning and data mining has been devoted to methods for learning ways to represent information in an unsupervised fashion, directly from observed data, aiming to lower the dimensionality of the observations, simplify them and uncover properties and structures of their latent driving system \cite{bengio2013representation}. This field of research has recently gained considerable attention due to the ability to acquire, store and share large amounts of information, leading to the availability of large scale data sets. Such large data sets are often generated by systems, which are not well understood and for which tailoring data representations based on prior knowledge is impossible. 

A particular representation learning subclass of methods is manifold learning \cite{bengio2013representation,tenenbaum2000global,coifman2006diffusion,belkin2001laplacian,scholkopf1997kernel,zhang2007mlle, roweis2000nonlinear}. In manifold learning it is assumed that observed data lie near a low-dimensional manifold embedded in a higher-dimensional ambient observation space. Typically the goal then is to embed the data in a low-dimensional Euclidean space while preserving some of their properties. 

Most manifold learning methods operate directly on the observed data and rely on their geometric properties \cite{tenenbaum2000global, bengio2013representation, coifman2006diffusion, donoho2003hessian, kamada1989algorithm, roweis2000nonlinear, saul2003think, singer2008non, tipping1999probabilistic}. This observed geometry, however, can be significantly influenced by the observation modality which is often arbitrary and without a clear known connection to the latent system of interest. Consequently, it is subpar to preserve observed properties, which might be irrelevant, and instead, one should seek to rely on intrinsic properties of the data, which are inherent to the latent system of interest and invariant to the observation modality.

Intrinsic geometry can be especially beneficial when it adheres to some useful structure, such as that of a vector space. Indeed there exist many systems whose dynamics and state space geometry can be simply described within an ambient Euclidean space. Unfortunately, when such systems are observed via an unknown non-linear observation function, this structure is typically lost. In such cases, it is advantageous to not only circumvent the influence of the observation function but also to explicitly preserve the global intrinsic geometry of the latent system.

We claim that in many situations, methods which are both intrinsic and globally isometric are better suited for manifold learning. In this paper, we propose a dimensionality reduction method which robustly estimates the push-forward metric of an unknown observation function using a parametric estimation implemented by an \ac{ANN}. The estimated metric is then used to calculate intrinsic-isometric geometric properties of the latent system state space directly from observed data. Our method uses these geometric properties as constraints for embedding the observed data into a low-dimensional Euclidean space. The proposed method is thus able to uncover the underlying geometric structure of a latent system from its observations without explicit knowledge of the observation model. 

The structure of this paper is as follows. In \cref{sec:toy_problem_and_motivation} we present the general problem of localization based on measurements acquired using an unknown observation model. This problem provides motivation for developing an intrinsic-isometric manifold learning method. In \cref{sec:problem_formulation} we formulate the problem mathematically. In \cref{sec:intrinsic_isometric_manifold_learning} we present our dimensionality reduction method, which makes use of the push-forward metric as an intrinsic metric. In \cref{sec:intrinsic_metric_estimation}, we propose a robust method for estimating this intrinsic metric directly from the observed data using an \ac{ANN}. In \cref{sec:results} we present the results of our proposed algorithm on synthetic data sets. In addition, we revisit the localization problem described in \cref{sec:toy_problem_and_motivation} in the specific case of image data and show that the proposed intrinsic-isometric manifold learning approach indeed allows for sensor invariant positioning and mapping in realistic conditions using highly non-linear observation modalities. In \cref{sec:conclusions} we conclude the work, discuss some key issues and outline possible future research directions.

\section{The localization problem and motivation}
\label{sec:toy_problem_and_motivation}

\begin{figure}[h]
		\begin{centering}
			\includegraphics[width=0.7\textwidth]{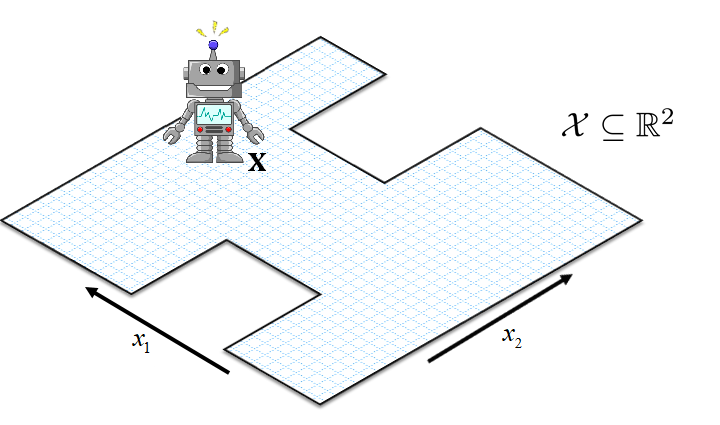}
			\par\end{centering}
		\caption{Agent in a $2$-dimensional space\label{fig:An-agent-in}}
	\end{figure}

Our motivating example is that of mapping a 2-dimensional region and positioning an agent within that region based on measurements it acquires. We denote the set of points belonging to the region by $\mathcal{X}\subseteq\mathbb{R}^{2}$ and the location of the agent by $\mathbf{x}\in\mathcal{X}$ (as illustrated in \cref{fig:An-agent-in}). At each position the agent makes measurements $\mathbf{y}=f(\mathbf{x})$, which are functions of its location. The values of the measurements are observable to us; however, we cannot directly observe the shape of $\mathcal{X}$ or the location of the agent $\mathbf{x}$, therefore they represent a latent system of interest.
	
	\begin{figure}[h]
		\begin{centering}
			\includegraphics[width=0.7\textwidth]{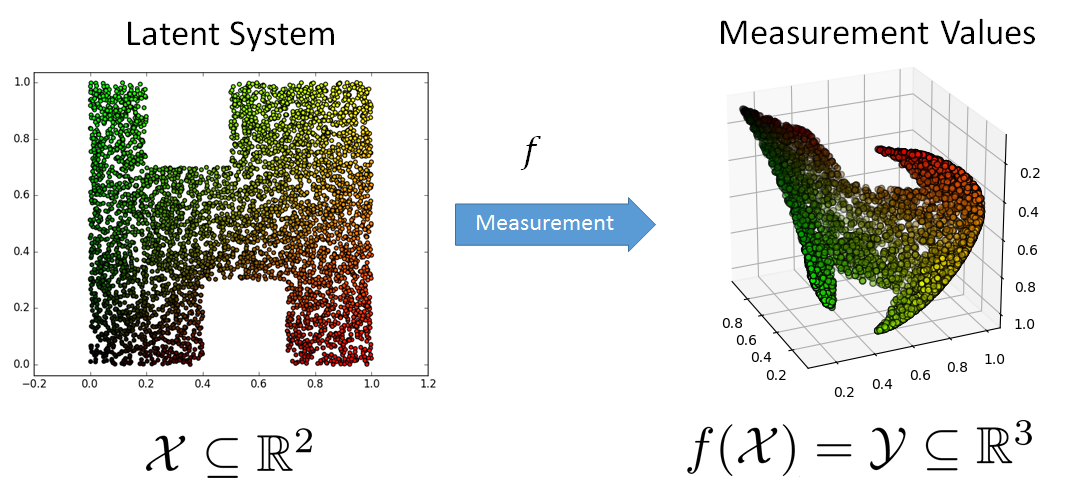}
			\par\end{centering}
		\caption{Creation of the observed manifold\label{fig:Creation-of-observed}}
	\end{figure}

\begin{figure}
		\centering
		\begin{subfigure}{.5\textwidth}
			\centering
			\includegraphics[width=.5\linewidth]{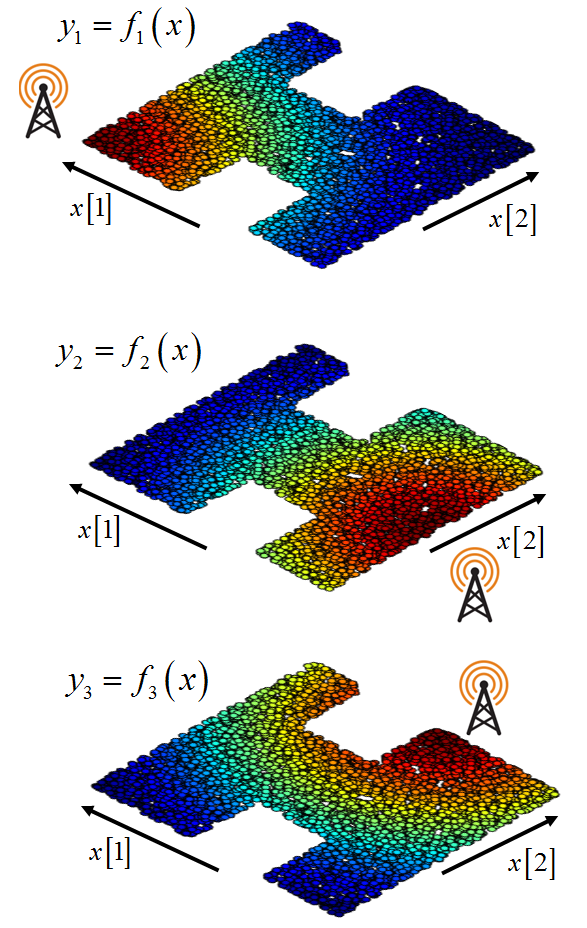}
			\caption{First measurement modality\label{fig:First-measurement-modality}}
		\end{subfigure}%
		\begin{subfigure}{.5\textwidth}
			\centering
			\includegraphics[width=.5\linewidth]{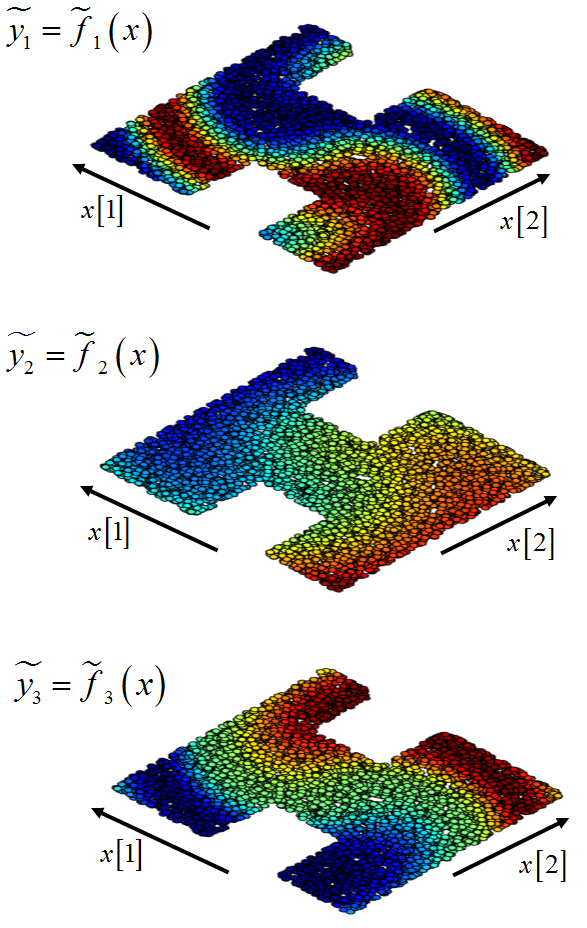}
			\caption{Second measurement modality\label{fig:Second-measurement-modality}}
		\end{subfigure}
			\caption{Two different observation function modalities\label{fig:test}. \protect\subref{fig:First-measurement-modality} A ``free-space'' signal strength decay model. The antenna symbol represents the location from which the signal originates \protect\subref{fig:Second-measurement-modality} A second measurement modality of arbitrary and unknown nature. 
			In both cases the variation in color represents different measurement values}. 
	\end{figure}

We simultaneously consider two distinct possible measurement models as described in \cref{fig:test}. The first measurement modality described in \cref{fig:First-measurement-modality} is conceptually similar to measuring \ac{RSS} values at $\mathbf{x}$ from antennas transmitting from different locations. Such measurements typically decay as the agent is further away from the signal origin. The second possible measurement model, visualized in \cref{fig:Second-measurement-modality}, consists of measurements, which are more complex and do not have an obvious interpretable connection to the location of the agent. Although both of these measurement modalities are 3-dimensional, the set of observable measurements resides on a 2-dimensional manifold embedded in 3-dimensional ambient observation space, as visualized in \cref{fig:Creation-of-observed}. This is due to the inherit 2-dimensional nature of the latent system.
	
The respective observation functions, $f$ and $\widetilde{f}$, of these measurement modalities are both bijective on $\mathcal{X}$, i.e. no two locations within $\mathcal{X}$ correspond to the same set of measurement values. Therefore, one can ask whether the location of the agent $\mathbf{x}$ can be inferred given the corresponding measurement values. If the measurement model represented by the observation function is known, this becomes a standard problem of model-based localization, which is a specific case of non-linear inverse problems \cite{engl2005nonlinear}. However, if the observation function is unknown or depends on many unknown parameters, recovering the location from the measurements becomes much more challenging. The latter case represents many real-life scenarios. For example, acoustic and electromagnetic wave propagation models, which are complex and depends on many a priori unknown factors such as room geometry, locations of walls, reflection, transmission, absorption of materials, etc. Image data can also serve as a possible indirect measurement of position. we note that although the image acquisition model is well known, the actual image acquired also depends on the geometry and look of the surrounding environment. If the surrounding environment is not known a priori then this observation modality also falls into the category of observation via an unknown model. 

In this work, we will consider the case of observation via an unknown model, and address the following question: is it be possible to retrieve $\mathbf{x}$ from $f\mathbf{\left(\mathbf{x}\right)}$ without knowing $f$? 

\begin{figure}[h]
		\centering{}%
		\begin{minipage}[t]{1\columnwidth}%
			\begin{center}
				\includegraphics[width=1\textwidth]{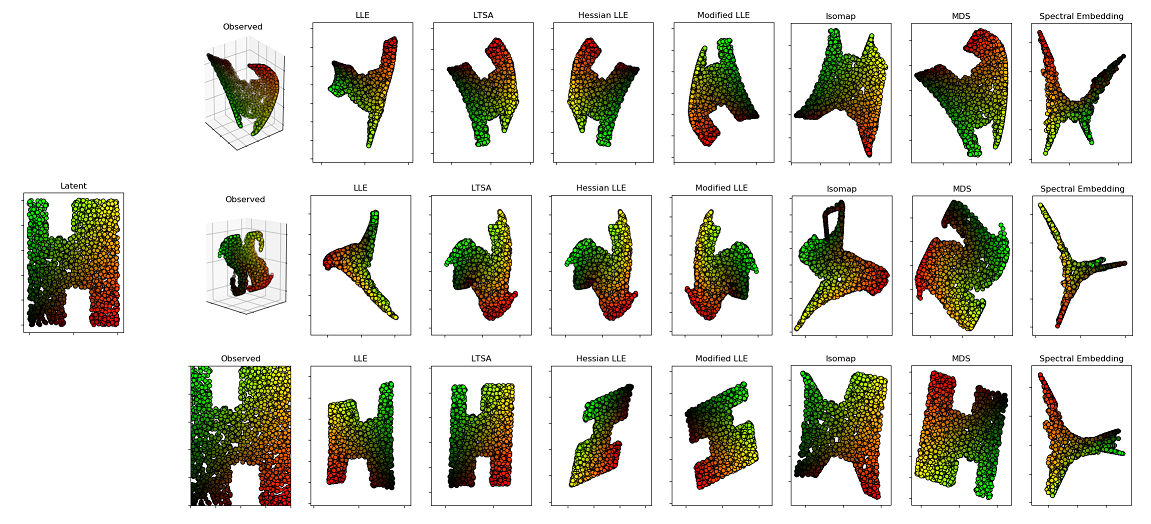}
				\par\end{center}%
		\end{minipage}\caption{Manifold learning for positioning and mapping based on observed measurements \label{fig:Application-of-manifold}. The first row corresponds to the application of manifold learning methods to the manifold created via the observation function visualized in \cref{fig:First-measurement-modality}. The second row corresponds to application of manifold learning methods to the manifold created via the observation function visualized in \cref{fig:Second-measurement-modality}. The bottom row represents application of manifold learning methods directly to the latent manifold}
	\end{figure}

Given that the described problem involves the recovery of a low-dimensional structure from an observable manifold embedded in a higher-dimensional ambient space, we are  inclined to use a manifold learning approach. However, standard manifold learning algorithms do not yield satisfactory results, as can be seen in \cref{fig:Application-of-manifold}. While some of the methods provide an interpretable low-dimensional parametrization of the latent space by preserving the basic topology of the manifold, none of the tested methods recover the true location of the agent or the structure of the region it moves in.
	
This problem demonstrates the inadequacy of existing manifold learning algorithms for uncovering the intrinsic geometric structure of a latent system of interest from its observations. We attribute this inadequacy to two main factors: the lack of \textit{intrinsicness} and the lack of \textit{geometry preservation} or \textit{isometry}.
Existing manifold learning algorithms generate new representations of data, while preserving certain \textit{observed} properties. When data originates from a latent system measured via an observation function, the observed measurements are affected by the specific (and often arbitrary) observation function, which in turn affects the learned low-dimensional representation. As a consequence, the same latent system observed via different observation modalities is represented differently when manifold learning methods are applied (as can be seen in \cref{fig:Application-of-manifold}). This fails to capture the intrinsic characteristics of the different observed manifolds, which originate from the same latent low-dimensional system. 
Settings involving different measurement modalities of unknown measurement modalities necessitate manifold learning algorithms, which are \textit{intrinsic}, i.e., that their learned representation is independent of the observation function and invariant to the way in which the latent system is measured.
	
Intrinsicness by itself is not enough in order to retrieve the latent geometric structure of the data.  be seen in \cref{fig:Application-of-manifold}, even when the same manifold learning methods are applied directly to the low-dimensional latent space, thus avoiding any observation function, preservation of the geometric structure of the latent low-dimensional space is not guaranteed. In order to explicitly preserve the geometric structure of the latent manifold, the learned representation needs to also be \textit{isometric}, i.e., distance preserving.

	
In this paper, we present a manifold learning method, which is \textit{isometric with respect to the latent intrinsic geometric structure}. We will show that, under certain assumptions, our method allows for the retrieval of $\mathbf{x}$ and $\mathcal{X}$ from observed data without requiring explicit knowledge of the specific observation model, thus solving a ``blind inverse problem''. In the context of motivating positioning problem, we achieve sensor-invariant localization, enabling the same positioning algorithm to be applied to problems from a broad range of signal domains with multiple types of sensors.
	
\section{Problem formulation}
\label{sec:problem_formulation}

Let ${\cal X}$ be a path-connected $n$-dimensional manifold embedded in $\mathbb{R}^{n}$. We refer to ${\cal X}$ as the intrinsic manifold and it represents the set of all the possible latent states of a low-dimensional system. These states are only indirectly observable via an observation function $f:{\cal X}\rightarrow\mathbb{R}^{m}$ which is a continuously-differentiable injective function that maps the latent manifold into the observation space $\mathbb{R}^{m}$. The image of $f$ is denoted by ${\cal Y}=f\left({\cal X}\right)$ and is referred to as the observed manifold. Since $f$ is injective we have that $n \leq m$. 

Here, we focus on a discrete setting, and define two finite sets of sample points from $\mathcal{X}$ and $\mathcal{Y}$. Let ${\cal X}_{s}=\left\{ {\bf x}_{i}\right\} _{i=1}^{N}$ be the set of $N$ intrinsic points sampled from $\mathcal{X}$, and let ${\cal Y}_{s}=f\left({\cal X}_{s}\right)$ be the corresponding set of observations of ${\cal X}_{s}$.
	
Under the above setting and given access only to $\mathcal{Y_{s}}$, we wish to generate a new embedding ${\cal \widetilde{X}}_{s}=\left\{ \tilde{{\bf x}}_{i}\right\} _{i=1}^{N}$ of the observed points ${\cal Y}_{s}=\left\{ {\bf y}_{i}\right\} _{i=1}^{N}$ into $n$-dimensional Euclidean space while respecting the geometric structure of the latent sample set $\mathcal{X}_{s}$. 

In order to provide a quantitative measure for ``intrinsic structure preservation'', we utilize the stress function commonly used for \ac{MDS} \cite{borg2005modern,de2005applications, buja2008data}. This function penalizes the discrepancies between the pairwise distances in the constructed embedding and an ``ideal'' target distance or dissimilarity. In our case the target distances are the true pairwise distances in the intrinsic space. This results in the following cost function:

\begin{equation}
	\sigma\left({\cal \widetilde{X}}_{s}\right)=\sum_{i<j}\left(\left\Vert \mathbf{\widetilde{x}}_{i}-\mathbf{\widetilde{x}}_{j}\right\Vert -\left\Vert {\bf x}_{i}-{\bf x}_{j}\right\Vert \right)^{2}
	\label{eq:stress}
\end{equation}
Namely, low stress values imply that the geometric structure of the embedding respects the intrinsic geometry conveyed by the pairwise distances in $\mathcal{X}$. Our goal then is to find a set of embedded points ${\cal \widetilde{X}}_{s}=\left\{ \tilde{{\bf x}}_{i}\right\} _{i=1}^{N}$ that minimizes $\sigma_{stress}\left({\cal \widetilde{X}}_{s}\right)$. 
	
Minimizing $\sigma_{stress}\left({\cal \widetilde{X}}_{s}\right)$ is seemingly a standard \ac{LS-MDS} problem, which is considered challenging, since it leads to a non-convex optimization problem \cite{de2005applications, gansner2004graph}, typically requiring a ``good'' initialization point located within the main basin of attraction, and hence, facilitating convergence to the global minimum. However, a significant additional challenge in our specific task is approximating the ground-truth pairwise Euclidean distances $\left\Vert {\bf x}_{i}-{\bf x}_{j}\right\Vert $ from the observed data even though the observation function $f$ is unknown.

	\externaldocument{paper.tex}

\section{Intrinsic isometric manifold learning}
\label{sec:intrinsic_isometric_manifold_learning}
	
In this section we describe a manifold learning algorithm, which is both intrinsic and isometric. 
		
Let $\mathbf{M}\left({\bf y}_{i}\right)$ denote the push-forward metric tensor on $\mathcal{Y}$ with respect to the diffeomorphism between $\mathcal{X}$ and $\mathcal{Y}$ defined by:
\begin{equation}
	\mathbf{M}\left({\bf y}_{i}\right)=\left[\frac{df}{dx}\left(f^{-1}\left({\bf y}_{i}\right)\right)\frac{df}{dx}\left(f^{-1}\left({\bf y}_{i}\right)\right)^{T}\right]^{\dagger}
\end{equation}
where $\frac{df}{dx}\left(f^{-1}\left({\bf y}_{i}\right)\right)$ is the Jacobian of the observation function $f$ with respect to the intrinsic variable at the intrinsic point ${\bf x}_i = f^{-1}\left({\bf y}_{i}\right)$.
The algorithm requires, as a prerequisite, that $\mathbf{M}\left({\bf y}_{i}\right)$ be known for all ${\bf y}_{i}\in{\cal Y}_{s}$. This appears to be a restrictive condition that requires knowledge of the observation function $f$. Nevertheless, we show in \cref{sec:intrinsic_metric_estimation} that in several scenarios these metrics can be robustly approximated from the accessible sample set ${\cal Y}_{s}$ without explicit knowledge of $f$.

\subsection{Local intrinsic geometry approximation}
\label{ssec:local_intrinsic_geometry_approximation}
		
The minimization of the cost function given in \eqref{eq:stress} first requires the approximation of the intrinsic Euclidean pairwise distances $d_{i,j}=d_{Euclidean}\left(\text{\ensuremath{\mathbf{x}}}_{i},\text{\ensuremath{\mathbf{x}}}_{j}\right)=\left\Vert {\bf x}_{i}-{\bf x}_{j}\right\Vert $ without direct access to the latent intrinsic data points $\mathcal{X}_s$. We overcome this problem by approximating these distances using the push-forward metric as follows:
\begin{equation}
d_{i,j}^{2}=\tilde{d}_{i,j}^2+\ensuremath{\mathcal{O}}\left(\left\Vert \text{\ensuremath{\mathbf{y}}}_{i}-\mathbf{y}_{j}\right\Vert ^{4}\right)
\label{eq:approx_plus_error}
\end{equation}
where the approximation $\tilde{d}_{i,j}$ is given by
\begin{equation}
	\tilde{d}_{i,j}^2 = \frac{1}{2}\left[\text{\ensuremath{\mathbf{y}}}_{i}-\mathbf{y}_{j}\right]^{T}\mathbf{M}\left({\bf y}_{i}\right)\left[\text{\ensuremath{\mathbf{y}}}_{i}-\mathbf{y}_{j}\right]+\frac{1}{2}\left[\text{\ensuremath{\mathbf{y}}}_{i}-\mathbf{y}_{j}\right]^{T}\mathbf{M}\left({\bf y}_{j}\right)\left[\text{\ensuremath{\mathbf{y}}}_{i}-\mathbf{y}_{j}\right]
\label{eq:int_dist_approx}
\end{equation}
This approximation has been used is several papers \cite{dsilva2013nonlinear,dsilva2015data,dsilva2015parsimonious,duncan2013identifying,mishne2015graph,singer2008non,talmon2012parametrization,talmon2013empirical,talmon2015intrinsic,talmon2015manifold} and error analysis for it was presented in \cite{dsilva2015data,singer2008non}. The approximation can be viewed as a result of the relation between the intrinsic and observed manifolds. If one recognizes that ${\cal X}$ can be viewed as a Riemannian manifold then $\mathbf{M}\left({\bf y}_{i}\right)$ is the push-forward metric tensor on $\mathcal{Y}$ with respect to the diffeomorphism between $\mathcal{X}$ and $\mathcal{Y}$. By using this metric on $\mathcal{Y}$, an isomorphism is established between the two manifolds $\mathcal{X}$ and $\mathcal{Y}$ which enables us to use $\mathcal{Y}$ as a proxy for making intrinsic geometric calculations on the latent manifold $\mathcal{X}$ just as if they were calculated directly on the latent intrinsic manifold $\mathcal{X}$. This effectively ``ignores" the observation function $f$ and the distortions it induces on the geometry of the observed manifold. 

Since the push-forward metric changes at each point on the manifold, this distance approximation is only locally valid. The locality of the approximation is demonstrated empirically in \cref{sec:results}. In addition, rigorous error analysis was provided in \cite{dsilva2015data}, where it was shown that the approximation is accurate for points on the observed manifold when $\left\Vert\text{\ensuremath{\mathbf{y}}}_{i}-\mathbf{y}_{j}\right\Vert ^{4}$ is sufficiently small with respect to the higher derivatives of the observation function.

\subsection{Embedding via partial-stress minimization}
\label{ssec:Embedding_construction_via_partial_stress_minimization}
		
Once the local pairwise intrinsic Euclidean distances are estimated, our goal then is to build an embedding of the observed manifold into $n$-dimensional Euclidean space, which respects those distances. However, due to the local nature of the distance estimation we only have valid approximation of short intrinsic distances and we cannot directly minimize the ``stress'' function presented in \eqref{eq:stress}. Instead we use the following weighted variant of the stress function:

\begin{equation}
\sigma_{w}\left({\cal \widetilde{X}}_{s}\right)=\sum_{i<j}w_{i,j}\left(\left\Vert \mathbf{\widetilde{x}}_{i}-\mathbf{\widetilde{x}}_{j}\right\Vert -\tilde{d_{i,j}}\right)^{2}\label{eq:w-intrinsic_stress}
\end{equation}
where unknown (or unreliable) distance estimations are assigned with zero weight and are thus excluded from the construction of the low-dimensional embedding. 

Since the observation function is unknown one cannot calculate the error term in the approximation \eqref{eq:approx_plus_error}. As a result, determining the scale at which distance estimations are deemed reliable from the observed data itself is still an open issue which is similar to local scale determination issues for many other common non-linear manifold learning methods. A good rule of thumb is to select for each point its distance to its $k$ nearest neighbors, where for most datasets tested, the use of 10-30 nearest neighbors provided the best results.

While the full stress function \eqref{eq:stress} can be globally optimized via eigenvector decomposition, this can no longer be done for \eqref{eq:w-intrinsic_stress}. However, it can be minimized via an iterative descent algorithm where the partial stress function is repeatedly upper bounded by a simple convex function whose exact minimum can be analytically found. This optimization method is referred to as \ac{SMACOF} optimization \cite{de2005applications, gansner2004graph} and it was adapted to the weighted stress function described in \eqref{eq:w-intrinsic_stress}.
			
Since the cost function \eqref{eq:w-intrinsic_stress} is non-convex with respect to the embedding coordinates, the iterative optimization process is only guaranteed to converge to some local minimum. In order to converge to a ``good'' minimum, the optimization process requires a good initialization point, that represents a possible embedding of the data into low-dimensional Euclidean space which is roughly similar to the actual low-dimensional intrinsic structure.

\subsection{Embedding initialization using intrinsic Isomap}
\label{ssec:Embedding_initialization_using_intrinsic_Isomap}

In order to construct an initial embedding for the \ac{LS-MDS} problem described in \cref{ssec:Embedding_construction_via_partial_stress_minimization}, we propose the use of an intrinsic variant of Isomap \cite{tenenbaum2000global}. This variant uses intrinsic, instead of observed, geodesic distances. These intrinsic geodesic distances can be approximated from the already calculated approximations of local intrinsic Euclidean distances by first constructing a graph from the data points, where only properly approximated intrinsic distance are represented by weighted edges, and then solving an ``all-to-all'' shortest path problem. The use of the push-forward metric and the shortest path algorithm provides a numerical approximation of the intrinsic geodesic distance. Once all intrinsic pairwise geodesic distances are approximated one can use classical \ac{MDS} to create an $n$-dimensional embedding in which Euclidean distances best respect the approximated geodesic distances. The main difference between this method and the standard Isomap algorithm is that instead of assuming local isometry (as is the case for Isomap) we use the push-forward metric on the observed manifold in order to explicitly guarantee local isometry. 

To understand why the push-forward metric and the shortest path algorithm provide an approximation of the intrinsic geodesic distances, we look at the definition of path length on the manifold. 
Let $\gamma\left(t\right)\in\cal{X}$ where $t\in\left[a,b\right]$ be an arbitrary path on the intrinsic manifold. The length of this path can be calculated by the following integral:
\[
L_{\mathcal{X}}\left[\gamma\right]=\int_{a}^{b}\left|\gamma\left(t\right)'\right|dt
\]
Using the push-forward metric, one can rewrite this integral in the observed space:
\[
L_{\mathcal{X}}\left[\gamma\right]=L_{\mathcal{Y},f_{PF}}\left[f\left(\gamma\right)\right]=\int_{f\left(a\right)}^{f\left(b\right)}\sqrt{\left({f\left(\gamma\left(t\right)\right)'}^{T} \left(\frac{df}{dx}\left({\bf x}_{i}\right)\frac{df}{dx}\left({\bf x}_{i}\right)^{T}\right)^{\dagger} f\left(\gamma\left(t\right)\right)'\right)}dt
\]	

Where the sub-script $f_{PF}$ indicates that the push-forward metric with respect to the observation function $f$ is used as a metric. This equation states that all length calculations for corresponding paths on the pair of diffeomorphic manifolds $\mathcal{X}$ and $\mathcal{Y}$ will be identical if one uses the push-forward metric on the observed manifold $\mathcal{Y}$. In this calculation the push-forward metric accounts for local stretching and contraction of the manifold due to the observation function, and changes the way we measure distances in order to compensate for this. This gives us a practical method to calculate intrinsic path lengths using only observed data. Using intrinsic curve length calculation one can calculate geodesic distances as the infimum intrinsic length over all paths in $\mathcal{Y}$ which connect $f\left({\bf x}_{i}\right)$ and $f\left({\bf x}_{j}\right)$ . Since only a final set of points sampled from this manifold is given, this minimization over all continuous paths can be approximated by a minimizing over all paths passing through the limited set of sample points. This presents geodesic distance approximation as a shortest-path problem where only short distances for which we have proper distance approximations are represented by edges on the graph. 

For the restricted class of convex latent intrinsic manifolds, the Euclidean and geodesic metrics coincide \cite{rosman2010nonlinear} allowing the intrinsic variant of Isomap, which uses intrinsic geodesic distances, to directly recover a proper embedding of the points into a low-dimensional space. This can also be observed in our results when applying the algorithm to convex problems. However, many problems of interest have a non-convex intrinsic state-space for which the discrepancy between the two metrics results in a distorted embedding which will not properly respect the intrinsic Euclidean geometry \cite{rosman2010nonlinear}. For intrinsic manifolds which are not extremely non-convex, the discrepancy between geodesic and Euclidean intrinsic distances can be small and the resulting embedding produced is often ``quite close'' to a proper intrinsic isometric embedding. This suggests that it possible to use the embedding generated by intrinsic intrinsic variant of Isomap as a initial point for further iterative optimization of the partial stress function \cref{eq:w-intrinsic_stress}. 

This approach harnesses the ability of eigen-decomposition to give a globally optimal solution for an embedding problem given all pairwise distances and the ability of iterative optimization to use a partial stress function to only consider local intrinsic Euclidean structure. By combining the two we get the benefits of a globally optimal solution of an approximate problem and the ability to solve the exact problem by locally optimal methods

\begin{figure}[h]
	\begin{centering}
		\includegraphics[width=1\textwidth]{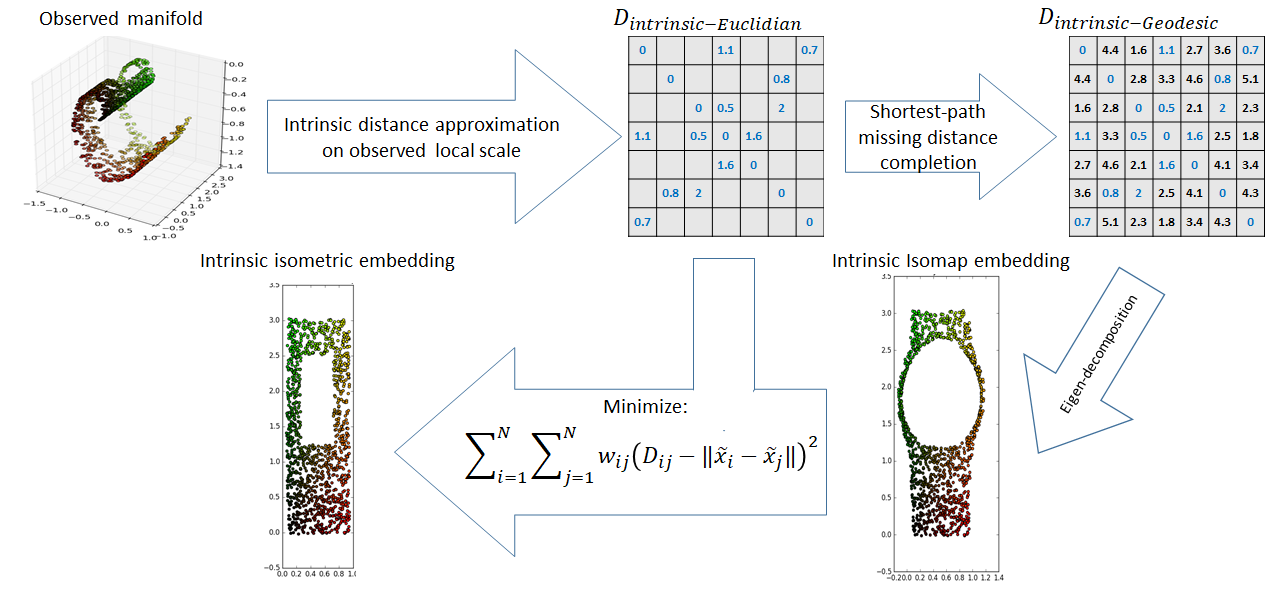}
		\par\end{centering}
	\caption{Flow graph of our intrinsic-isometric manifold learning method applied to an observed manifold} \label{fig:Intrinsic-isometric-manifold-lea}
\end{figure}

The flow of this algorithm is illustrated in \cref{fig:Intrinsic-isometric-manifold-lea}. 

\subsection{Failure detection and a multi-scale scheme}
\label{ssec:Multi_scale_scheme}

Convergence of the described iterative optimization process to a low-stress embedding is achieved, provided that the initial embedding obtained by our intrinsic variant of the Isomap algorithm is sufficiently close to the true intrinsic structure and that we have enough accurate pairwise distance estimates such the embedding problem is sufficiently constrained and well-posed as described in \cref{ssec:Embedding_construction_via_partial_stress_minimization}. 

The described embedding method could fail due to various adverse conditions such as extremely noisy samples, insufficient data, sparse sampling, non-convexity of the intrinsic manifold, etc. A poor intrinsic isometric embedding can be detected by lack of consistency between the intrinsic geodesic distances estimated from the observed data and the geodesic distances in the constructed low-dimensional embedding.

One of the possible causes of a failed embedding is a highly non-convex latent manifolds. To handle such manifolds we also devise an iterative multi-scale operation scheme aimed to take advantage of the fact that small patches of the manifold are typically convex or approximately convex, thus increasing the chance of the \ac{LS-MDS} optimization converging to a good solution. In this multi-scale scheme, large patches of points on the manifold are split to partially overlapping patches. Each of the sub-patches can be further recursively split to smaller and smaller patches until small enough patches, that are not extremely non-convex and allow for the successful application of our algorithm, ,are reached.

Once two sub-patches are successfully embedded into a low dimensional Euclidean space they are registered using a rotation transformation which optimally aligns their overlapping points. After registration, these patches are merged into a larger scale patch. To avoid the accumulation of error stemming from this bottom-up procedure, the embedding obtained by merging of the embedding of the sub-patches is used only as an initial embedding that is further iteratively optimized as described in \cref{ssec:Embedding_initialization_using_intrinsic_Isomap}.

\begin{figure}[h]
	\begin{centering}
		\begin{minipage}[c]{.19\textwidth}
			\begin{subfigure}[t]{1\textwidth}%
				\includegraphics[width=1\textwidth]{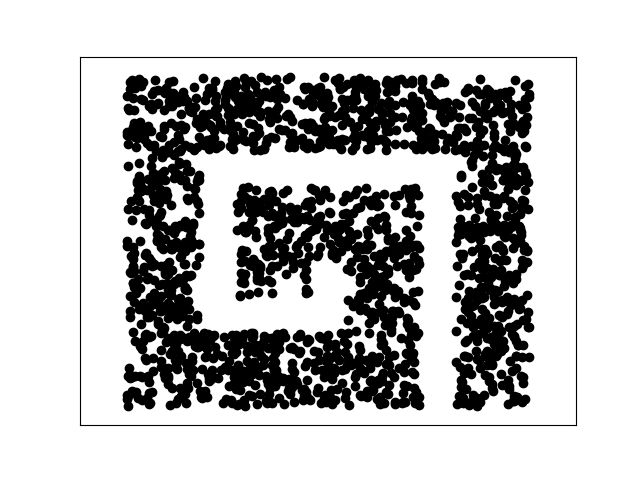}
				\caption{\label{fig:multi_a}}
			\end{subfigure}
		\end{minipage}
		\begin{minipage}[c]{.19\textwidth}
			\begin{subfigure}[t]{1\textwidth}%
				\includegraphics[width=1\textwidth]{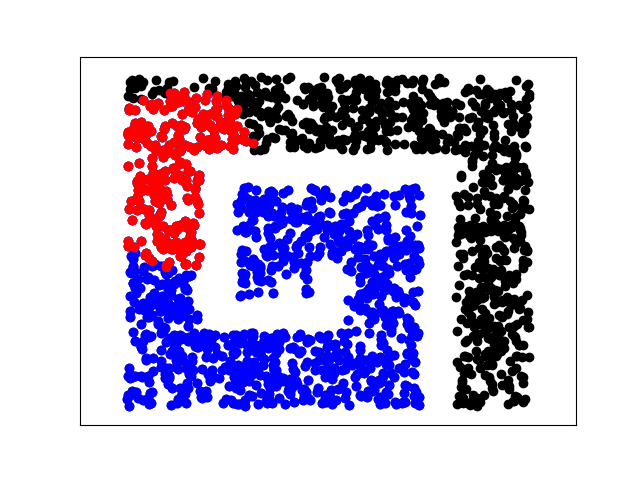}
				\caption{\label{fig:multi_b}}
			\end{subfigure}
			\begin{subfigure}[t]{1\textwidth}%
				\includegraphics[width=1\textwidth]{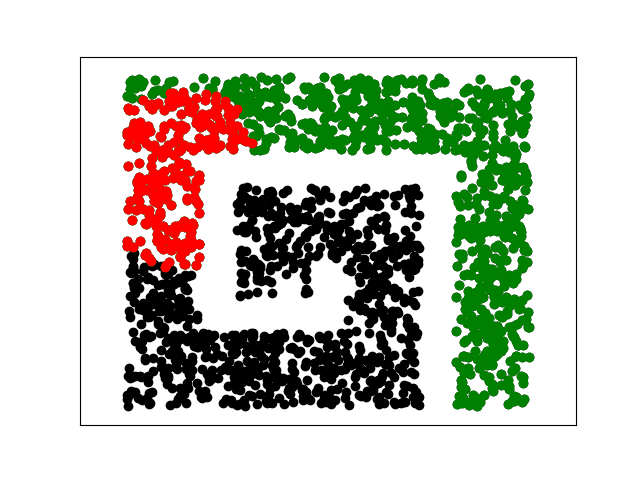}
				\caption{\label{fig:multi_c}}
			\end{subfigure}\hfill
		\end{minipage}
		\begin{minipage}[c]{.19\textwidth}
			\begin{subfigure}[t]{1\textwidth}%
				\includegraphics[width=1\textwidth]{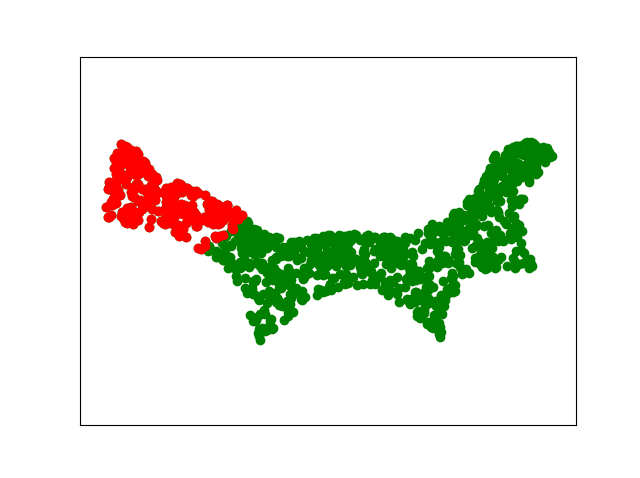}
				\caption{\label{fig:multi_d}}
			\end{subfigure}
			\begin{subfigure}[t]{1\textwidth}%
				\includegraphics[width=1\textwidth]{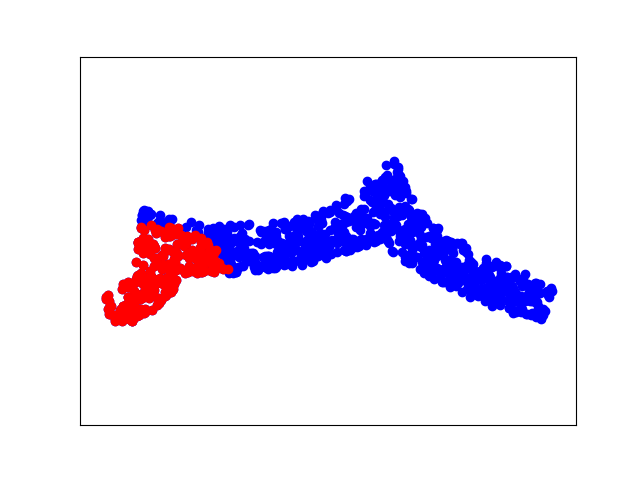}
				\caption{\label{fig:multi_e}}
			\end{subfigure}\hfill
		\end{minipage}
		\begin{minipage}[c]{.19\textwidth}
			\begin{subfigure}[t]{1\textwidth}%
				\includegraphics[width=1\textwidth]{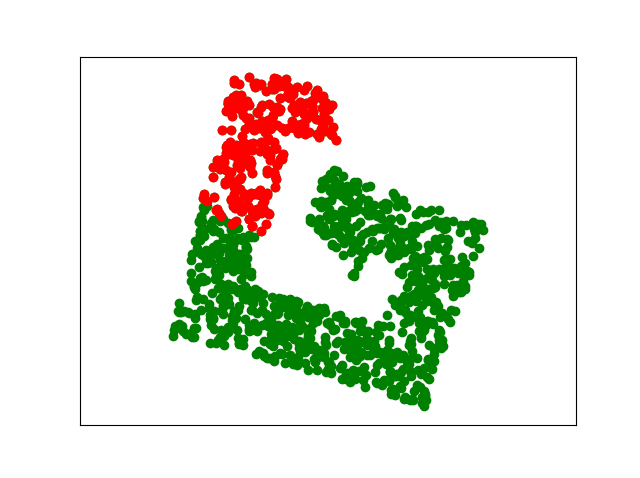}
				\caption{\label{fig:multi_f}}
			\end{subfigure}
			\begin{subfigure}[t]{1\textwidth}%
				\includegraphics[width=1\textwidth]{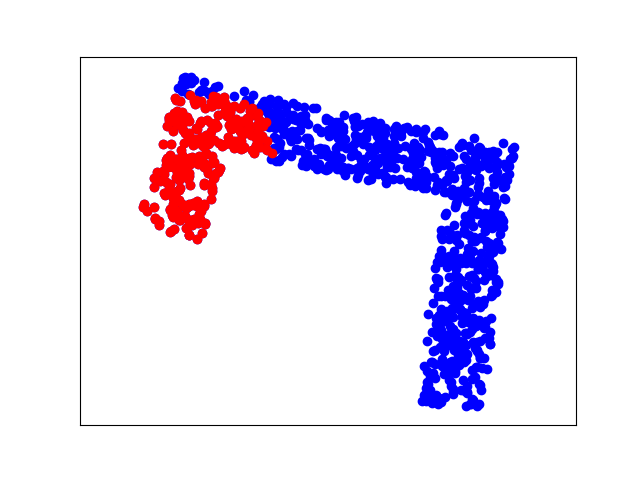}
				\caption{\label{fig:multi_g}}
			\end{subfigure}\hfill
		\end{minipage}
		\begin{minipage}[c]{.19\textwidth}
			\begin{subfigure}[t]{1\textwidth}%
				\includegraphics[width=1\textwidth]{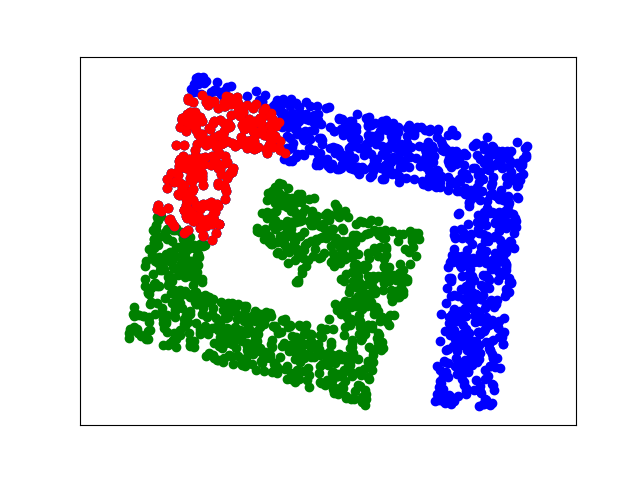}
				\caption{\label{fig:multi_i}}
			\end{subfigure}\hfill
		\end{minipage}
	\end{centering}
	\caption[Bottom up multiscale schema] {A highly non-convex manifold \protect\subref{fig:multi_a} is split to two partially overlapping sub-patches colored blue and green \protect\subref{fig:multi_b},\protect, \subref{fig:multi_c}. Their overlap is colored red. The initial embeddings of the sub-patches obtained by two separate applications of the intrinsic Isomap \protect\subref{fig:multi_d},\protect\subref{fig:multi_e}. Based on these two initial embeddings, intrinsic-isometric embeddings are constructed separately for each sub-patch \protect\subref{fig:multi_f},\protect\subref{fig:multi_g}, which are then registered and merged according to their overlapping areas to attain an embedding of the entire which is then further optimized using the LS-MDS optimization process.\label{fig:Multiscale}.

	\protect\subref{fig:multi_i} Final intrinsic isometric embedding of entire manifold.}
\end{figure}

The proposed multi-scale scheme is illustrated on a non-convex manifold depicted in \cref{fig:Multiscale}. We note that the splitting of the patches is implemented by using the already calculated estimates of the intrinsic geodesic distances between points. We further remark that while the visualization is in the intrinsic space, the estimation of the geodesic distances, and hence, the scale selection are performed directly on the observed manifold.

	\externaldocument{paper.tex}

\section{Intrinsic Metric Estimation}
\label{sec:intrinsic_metric_estimation}

For the algorithm presented in \cref{sec:intrinsic_isometric_manifold_learning} we assumed that the metric $\mathbf{M}\left({\bf y}_{i}\right)$ is given at every point $\text{\ensuremath{\mathbf{y}}}_{i}\in\mathcal{Y}_{s}$. This metric is the intrinsic ``push-forward'' metric, which allows us to approximate intrinsic Euclidean and geodesic distances and has a key role in recovering the intrinsic latent manifold. Therefore, in order to apply the algorithm proposed in \cref{sec:intrinsic_isometric_manifold_learning}, $\mathbf{M}\left({\bf y}_{i}\right)$, which is typically unknown, needs to be robustly approximated from the observed data. This approximation is the subject of this section.
	
	\subsection{Intrinsic isotropic Gaussian mixture model}
	\label{ssec:Intrinsic-isotropic-GMM}
	
	We present a basic setting, which allows the approximation of the intrinsic metric directly from the observed data. Assume that the intrinsic data are sampled from an $n$-dimensional \ac{GMM} consisting of $N$ isotropic Gaussian distributions, all with the same known variance $\sigma_{int}^{2}$, where the means of these distributions are points on the intrinsic manifold, denoted by $\left\{ \mathbf{x}_{i}\right\} _{i=1}^{N}=\mathcal{X}_{S}\subseteq\mathcal{X}$. To distinguish between the model means and the observed data points, the data points are denoted by double indexing, i.e., ${\bf x}_{i,j}\in\mathcal{X}$ and:
	\[
			{\bf x}_{i,j}\sim\mathcal{N}\left(\mathbf{x}_{i},\sigma_{int}^{2}\mathbf{I}\right), \ j=1,\ldots,N_i
	\]
	where $N_{i}$ is the number of points sampled from the $i$-th Gaussian and $\mathbf{I}$ is the identity matrix. The intrinsic points are observed via an observation function $f:\mathbb{R}^{n}\to\mathbb{R}^{m}$ and are subject to \ac{AWGN} with variance $\sigma_{obs}^{2}$ introduced in the observation process. This results in the following observation model:
			\[
			{\bf y}_{i,j}=f\left({\bf x}_{i,j}\right)+w_{i,j}
			\]
	where $w_{i,j}\sim\mathcal{N}\left(0,\sigma_{obs}^{2}\mathbf{I}\right)$. Broadly, if $\sigma_{int}^{2}$ is small with respect to the second derivative of $f$, the observed data follows a Gaussian distribution with transformed mean and covariance:
	\begin{equation}
			{\bf y}_{i,j}=f\left({\bf x}_{i,j}\right)+w_{i,j}\sim\mathcal{N}\left(f\left(\mathbf{x}_{i}\right),\sigma_{int}^{2}\frac{df}{dx}\left(\mathbf{x}_{i}\right)\frac{df}{dx}\left(\mathbf{x}_{i}\right)^{T}+\sigma_{obs}^{2}\mathbf{I}\right)\label{eq:prob-pca}
	\end{equation}
	For a more rigorous analysis, see \cite{dsilva2015data}. Namely, the observed data follows a \ac{GMM} distribution, with anisotropic covariance, induced by the Jacobian of the observation function. Each Gaussian in this \ac{GMM} can be interpreted according to probabilistic \ac{PCA}  \cite{tipping1999probabilistic}, where it is shown that the optimal maximum-likelihood estimator of the matrix $\frac{df}{dx}\left(\mathbf{x}_{i}\right)\frac{df}{dx}\left(\mathbf{x}_{i}\right)^{T}$ can be computed by applying \ac{PCA} and finding the $n$ most significant principal directions of the observed sample covariance matrix $\mathbf{S}_{i}$, i.e.:
		\[
			\mathbf{S}_{i} =\frac{1}{N_{i}}\sum_{j=1}^{N_{i}}\left[{\bf y}_{i,j}-{\bf \bar{y}}_{i}\right]\left[{\bf y}_{i,j}-{\bf \bar{y}}_{i}\right]^{T}
		\]
	where ${\bf \bar{y}}_{i}$ is the sample mean:
		\[
			{\bf \bar{y}}_{i} =\frac{1}{N_{i}}\sum_{j=1}^{N_{i}}{\bf y}_{i,j}
		\]
	This matrix can be decomposed using spectral decomposition:
			\[
				\mathbf{S}_{i}=\sum_{k=1}^{m}\sigma_{k}^{2}\mathbf{u}_{k}\mathbf{u}_{k}^{T}
			\]
	And the maximum-likelihood estimator is given by \cite{tipping1999probabilistic}:
		\begin{equation}
		\begin{aligned}
		\frac{df}{dx}\left(\mathbf{x}_{i}\right)\frac{df}{dx}\left(\mathbf{x}_{i}\right)^{T} & \approx
		\sum_{k=1}^{n}\left(\sigma_{k}^{2}-\bar{\sigma^{2}}\right)\mathbf{u}_{k}\mathbf{u}_{k}^{T}\\
		\bar{\sigma^{2}} & =\frac{1}{m-n}\sum_{k=n+1}^{m}\sigma_{k}^{2}
		\end{aligned}
		\end{equation}
	Assuming that the clustering structure of observed measurements is known (i.e. we are able to group together observation points originating from the same Gaussian component), $\mathbf{M}\left({\bf y}_{i}\right)=\frac{df}{dx}\left(\mathbf{x}_{i}\right)\frac{df}{dx}\left(\mathbf{x}_{i}\right)^{T}$ can be estimated directly from $\mathbf{S}_i$, separately for each model $i$. This setting is similar to the setting presented in \cite{singer2008non} where short bursts of intrinsic diffusion processes with isotropic diffusion were used, resulting in an approximate intrinsic \ac{GMM} model and leading to the same estimator of the intrinsic metric using local applications of \ac{PCA}.
			

		
	We note that a common characteristic of local estimation methods, such as the one described above, is that these methods typically operate at a scale on which the observation function $f$ is approximately linear and can be well approximated by its Jacobian. The curvature of the manifold dictates how fast the Jacobian and the metric changes along the manifold and therefore restricts the scale at which samples are relevant for estimation of the metric at a specific point. If the data is not sufficiently sampled, the number of data points that will be incorporated into the estimation of a single local metric will be small, resulting in an a poor estimation. This non-robustness is not unique to these specific metric estimation methods and many non-linear estimation problems and manifold learning techniques are non-robust due to their local nature as discussed in \cite{bengio2004non}. To overcome this limitation, the remainder of this section introduces global (parametric) regularization of the metric estimation, imposing a smooth variation of the estimated metrics along the intrinsic manifold.

	\subsection{Maximum-Likelihood intrinsic metric estimation}
	
	In \cref{ssec:Intrinsic-isotropic-GMM}, the observed data were modeled by a \ac{GMM} with covariance matrices that depend on the Jacobian of the observation function \cref{eq:prob-pca}. Under this statistical model, an unconstrained maximum-likelihood estimation of the metric $\mathbf{M}\left({\bf y}_{i}\right)$ is a local estimator, which only uses data from a single Gaussian and does not impose any smoothness on the metric, limiting the number of samples incorporated in the estimation and the estimation accuracy. 
	
	To overcome this, we propose a constrained non-local maximum-likelihood estimator; the intrinsic isotropic \ac{GMM} model described in \cref{ssec:Intrinsic-isotropic-GMM} results in the following approximate global log-likelihood function for the observed data:
	\begin{equation}
	\mathcal{L}\left(\left\{ \frac{df}{dx}\left(\mathbf{x}_{i}\right)\frac{df}{dx}\left(\mathbf{x}_{i}\right)^{T}\right\} _{i=1}^{N}\right)
	=
	-\frac{1}{2}\sum_{i=1}^{N}N_{i}\left\{ d\text{ln}\left(2\pi\right)+\text{ln}\left|\mathbf{C}_{i}\right|+\text{Tr}\left(\mathbf{C}_{i}^{-1}\mathbf{S}_{i}\right)\right\} 
	\end{equation}
	where $N$ is the number of Gaussians in the \ac{GMM}, $N_{i}$ is the number of points sampled from the $i$-th Gaussian, $\mathbf{S}_{i}$ is the sample covariance of the $i$-th observed Gaussian distribution and $\mathbf{C}_{i}$ denotes the (population) covariance which is given by:
	\begin{equation}
	\mathbf{C}_{i}=\sigma_{int}^{2}\frac{df}{dx}\left(\mathbf{x}_{i}\right)\frac{df}{dx}\left(\mathbf{x}_{i}\right)^{T}+\sigma_{obs}^{2}\mathbf{I}
	\end{equation}
	This log-likelihood is also presented in \cite{tipping1999mixtures}. 
	
	We restrict the Jacobian of the observation function to a parametric family of functions, which we denote by
	$\mathbf{J}\left({\bf y}_{i}|\theta\right):\mathbb{R}^{m}\to\mathbb{R}^{m\times n}$, where $\theta$ represents the parametrization of the family. 
	Estimation is then performed by choosing $\theta$ that maximizes the parameterized log-likelihood function:
	\begin{equation}
	\mathcal{L}\left(\left\{ {\bf y}_{i.j}\right\} |\theta\right)=-\frac{1}{2}\sum_{i=1}^{N}N_{i}\left\{ d\text{ln}\left(2\pi\right)+\text{ln}\left|{\bf C}\left({\bf y}_{i}|\theta\right)\right|+\text{Tr}\left(\mathbf{\mathbf{C}}\left({\bf y}_{i}|\theta\right)^{-1}\mathbf{S}_{i}\right)\right\} \label{eq:Ann-cost}
	\end{equation}
	where
	\[
	\mathbf{C}\left({\bf y}_{i}|\theta\right)=\sigma_{int}^{2}\mathbf{J}\left({\bf y}_{i}|\theta\right)\mathbf{J}\left({\bf y}_{i}|\theta\right)^{T}+\sigma_{obs}^{2}\mathbf{I}
	\]
	
	We note that the log-likelihood in \cref{eq:Ann-cost} only differs by a constant from the following function:
	\begin{equation}
	\begin{aligned}R\left(\left\{ {\bf y}_{i.j}\right\} |\theta\right) & =-\sum_{i=1}^{N}N_{i}D_{KL}\left(\mathbf{C}_{i}||\mathbf{S}_{i}\right)\\
	& =-\sum_{i=1}^{N}N_{i}\left\{ \text{ln}\frac{\left|\mathbf{C}_{i}\right|}{\left|\mathbf{S}_{i}\right|}+\text{Tr}\left(\mathbf{C}_{i}^{-1}\mathbf{S}_{i}\right)\right\} \\
	& =\mathcal{L}\left(\left\{ {\bf y}_{i.j}\right\} |\theta\right)+\frac{1}{2}\sum_{i=1}^{N}N_{i}\left\{ d\text{ln}\left(2\pi\right)-\text{ln}\left|\mathbf{S}_{i}\right|\right\} \\
	& =\mathcal{L}\left(\left\{ {\bf y}_{i.j}\right\} |\theta\right)+constant
	\end{aligned}
	\label{eq:KL}
	\end{equation}
	where $D_{KL}\left(\mathbf{C}_{i}||\mathbf{S}_{i}\right)$ is the well known Kullback\textendash Leibler divergence between Gaussian distributions with the same mean and two different covariances matrices $\text{\ensuremath{\mathbf{C}_{i}}}$ and $\mathbf{S}_{i}$. This divergence reaches its minimal value when $\text{\ensuremath{\mathbf{C}_{i}}}$ and $\mathbf{S}_{i}$ are equal. Thus maximization of the likelihood is equivalent to minimization of the Kullback\textendash Leibler divergence between the covariance, expressed by the parametric model, and the sample covariance, which is empirically built from data.

	\subsection{Parametric estimation using regularized artificial neural networks}
	
	To increase the robustness of the estimation, and specifically, to avoid large variations in the metric estimates, we use a parametric family of smooth functions, implemented by an \ac{ANN}. Neural networks are compositions of linear transformations and non-linear operation nodes, which can be adapted to approximate a specific function by choosing the values of the linear transformations in the network. These values are called the ``weights'' of the network. Thus, in the context of our work, the parameter vector $\theta$ consists of the weights of the network. By choosing the non-linear operation nodes to be smooth functions, e.g., the standard sigmoid function $s\left(x\right)=\frac{1}{1+e^{-x}}$, we can explicitly impose smoothness. 
	The input of the network are the observed data and the output are $n\times m$ matrices $\mathbf{J}\left({\bf y}_{i}|\theta\right)$, which are estimates of the Jacobian of the observation function at the input points. The estimation is carried out by optimizing the weights of the network using the likelihood \eqref{eq:Ann-cost} as the cost function.
	The neural network structure used in this work contains two hidden layers and an additional linear layer at the output, which was added in order to allow for automatic scaling of the output.

	The use of \acp{ANN} has several advantages with respect to other smooth parametric function families. First, \acp{ANN} are general and have proven to perform well in a wide spectrum of applications. Second, many methods for network regularization exist, facilitating convenient implementation of the smoothness constraint. Third, due to their extreme popularity, there are many software and hardware solutions allowing for efficient and fast construction and optimization.
	
	This estimation method also has some secondary advantages in addition to providing additional robustness to the intrinsic metric estimation when compared to local metric estimation methods. As opposed to local methods which only estimate the intrinsic metric for observed sample points, the regression approach gives rise to an estimation of the intrinsic metric on the whole observed space. This can be used to generate additional points in between existing points, thus artificially increasing the sample density. This can improve both the short-range intrinsic distance estimation described in \cref{ssec:local_intrinsic_geometry_approximation} and the geodesic distance estimation described in \cref{ssec:Embedding_construction_via_partial_stress_minimization}. Both effects could improve the results of the algorithm presented in this paper. In addition, it was shown in \cite{bengio2004non}, that by learning the tangent plane to the manifold at each point, one can ``walk'' on the manifold by making infinitesimal steps each time in the tangent plane. Since with the method proposed in this chapter, we do not only estimate the tangent plane but the intrinsic metric as well, we know how far we have gone in terms of the distance, an ability that might be relevant to applications such as non-linear interpolation \cite{bregler1995nonlinear}.

	\subsection{Implementation}
	\label{ssec:Net_implementation}
		
	In order to optimize the network weights with respect to non-standard cost function (presented in \cref{eq:Ann-cost}) we implemented the described metric estimation network in Python using Theano \cite{bergstra2010theano}, which allows for general symbolic derivation and gradient calculation. To facilitate better and faster optimization, the stochastic gradient descent method \ac{ADAM} \cite{kingma2014adam} was used. This method uses momentum and moment estimation over time to automatically and adaptively scale the learning rate of the optimization process for faster convergence. 
	
	We control the estimated function complexity by limiting the number of nodes in the network.
	In addition, we use a weight decay penalty term, which encourages small weights and, in general, prevents large variations of the output functions with respect to changes in the input \cite{krogh1991simple}. 
	
	In unsupervised settings, choosing the network hyper-parameters (e.g., the number of nodes and the emphasis put on the weight decay term) is not straight-forward, since there are no ``labels'' or ground truth values for the intrinsic metric (or the Jacobian). Here, it is carried out by cross-validation using the log-likelihood function as the cost function. In our tests, we used k-fold cross-validation where $20\%$ of the observed samples were used for validation purposes. This was repeated for each set of hyper-parameters, where the validation set was sampled randomly from the whole data at each iteration.

	\externaldocument{paper.tex}
\section{Experimental results} \label{sec:results}

	\subsection{Simulated data}
	
	Consider the intrinsic manifold $\mathcal{X}$ corresponding to a 2-dimensional square with a cross shaped hole at its center. This latent manifold is observed via the following observation function:
	\begin{equation} \label{eq:simulated_data_observation_function}
	\mathbf{y\left(\mathbf{x}\right)}=f\left(\mathbf{x}\right)=\text{\ensuremath{\left[\begin{array}{c}
			y_{1}\left(\mathbf{x}\right)\\
			y_{2}\left(\mathbf{x}\right)\\
			y_{3}\left(\mathbf{x}\right) 
			\end{array}\right]}=}
		\left[\begin{array}{c}
	\sin \left(2.5 \cdot x_1 \right)\cdot \sin \left( x_2 \right)\\
	\sin \left(2.5 \cdot x_1 \right)\cdot \cos \left( x_2 \right)\\
	-\sin \left( x_2 \right)
	\end{array}\right]
	\end{equation}
	The observation function embeds the data in 3-dimensional Euclidean space by wrapping it around the unit sphere resulting in the observed manifold $\mathcal{Y}=\left\{ f\left(\mathbf{x}\right) \mid \mathbf{x}\in\mathcal{X}\right\}$. The sample sets $\mathcal{X}_{s}$ and $\mathcal{Y}_{s}$ are generated by uniformly sampling $N=1000$ points from $\mathcal{X}$. 
	
	First we evaluate the validity of the intrinsic Euclidean distance estimation presented in \cref{eq:int_dist_approx} and the ability of the embedding method proposed in \cref{sec:intrinsic_isometric_manifold_learning} to recover the latent geometric structure of the data given the intrinsic metric over the observed manifold. To do so, we analytically calculate $\frac{df}{dx}\left(\mathbf{x}_{i}\right)\frac{df}{dx}\left(\mathbf{x}_{i}\right)^{T}$ by taking the derivative of the observation function given in \cref{eq:simulated_data_observation_function} with respect to the intrinsic latent variable. 
	
\begin{figure}[H]	
		\begin{centering}
			\begin{subfigure}[b]{0.3\linewidth}
				\includegraphics[width=1\linewidth]{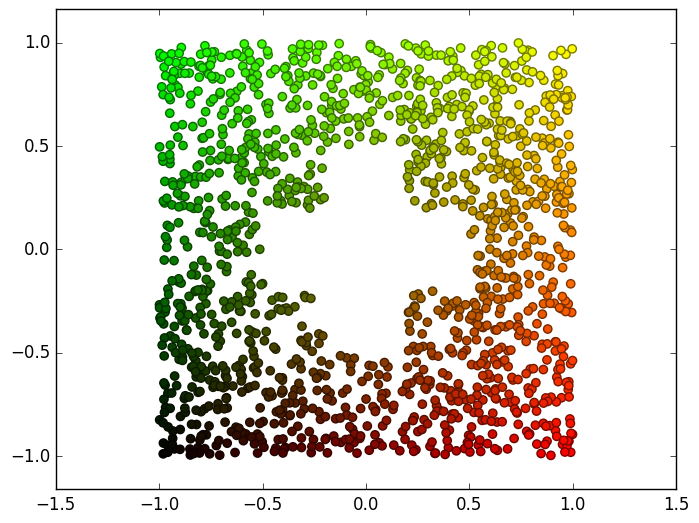}
				\caption{\label{fig:punctured_severed_sphere_intrinsic}}
			\end{subfigure}\hfill
			\begin{subfigure}[b]{0.3\linewidth}
				\includegraphics[width=1\linewidth]{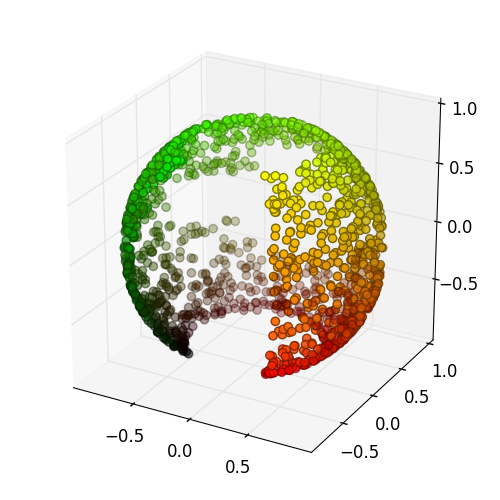}
				\caption{\label{fig:punctured_severed_sphere_observed}}
			\end{subfigure}\hfill
			\begin{subfigure}[b]{0.3\linewidth}
				\includegraphics[width=1\linewidth]{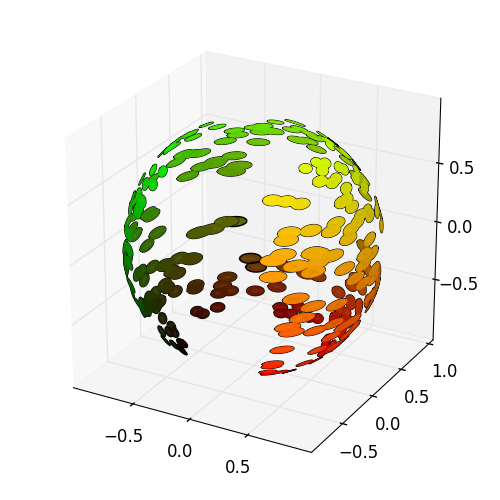}
				\caption{\label{fig:punctured_severed_sphere_intrinsic_metric}}
			\end{subfigure}
		\end{centering}
		\begin{centering}
			\begin{subfigure}[b]{0.45\linewidth}
				\includegraphics[width=1\linewidth]{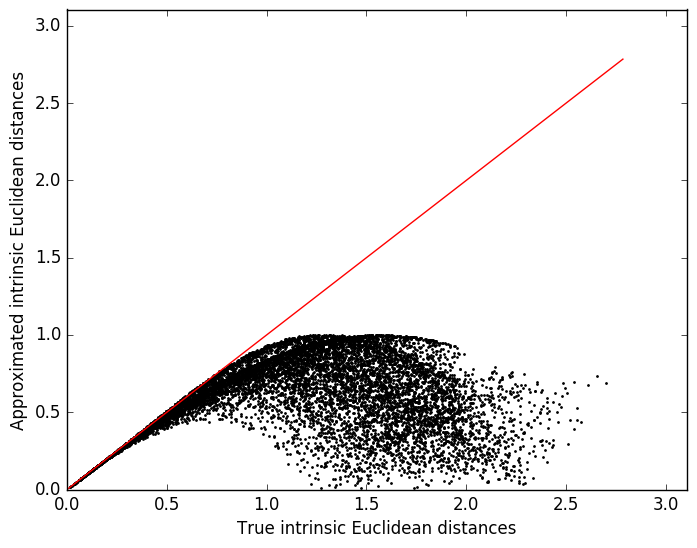}
				\caption{\label{fig:punctured_severed_sphere_intrinsic_dist_est}}
			\end{subfigure}
			\hfill
			\begin{subfigure}[b]{0.45\linewidth}
				\includegraphics[width=1\linewidth]{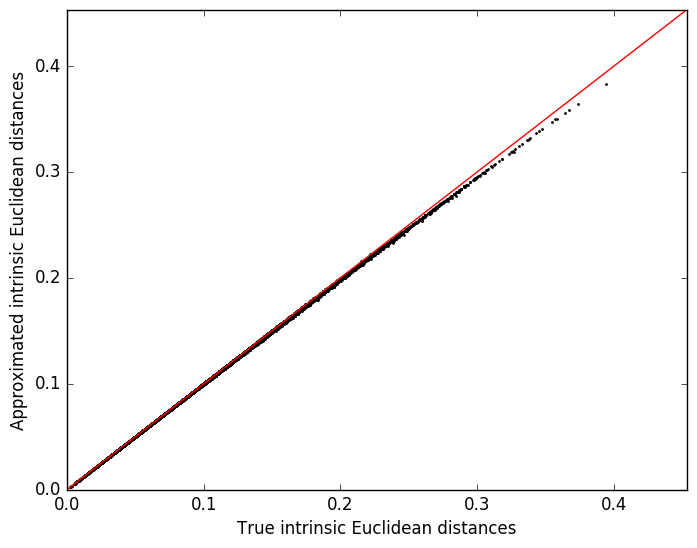}
				\caption{\label{fig:punctured_severed_sphere_intrinsic_dist_est_knn}}
			\end{subfigure}
		\end{centering}
		\begin{centering}
			\begin{subfigure}[b]{0.32\linewidth}
				\includegraphics[width=1\linewidth]{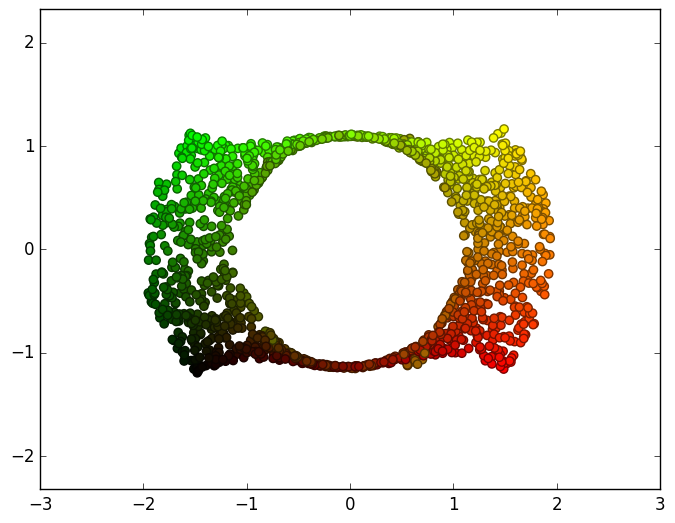}
				\caption{\label{fig:punctured_severed_sphere_standard_isomap_embedding}}
			\end{subfigure} \hfill
			\begin{subfigure}[b]{0.32\linewidth}
				\includegraphics[width=1\linewidth]{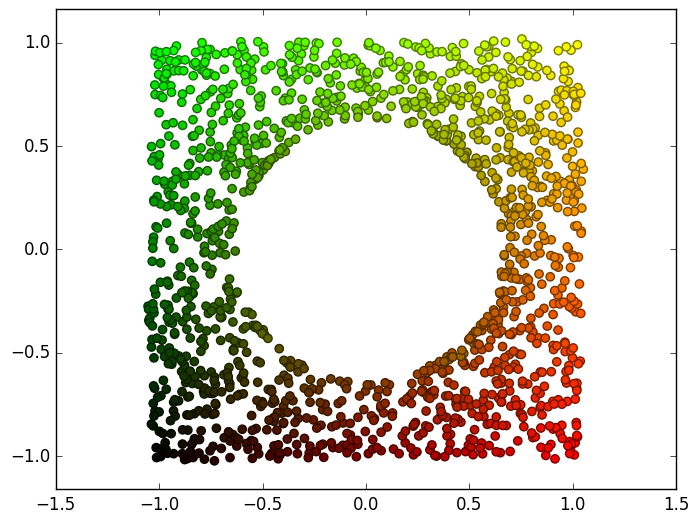}
				\caption{\label{fig:punctured_severed_sphere_intrinsic_isomap_embedding}}
			\end{subfigure} \hfill
			\begin{subfigure}[b]{0.32\linewidth}
				\includegraphics[width=1\linewidth]{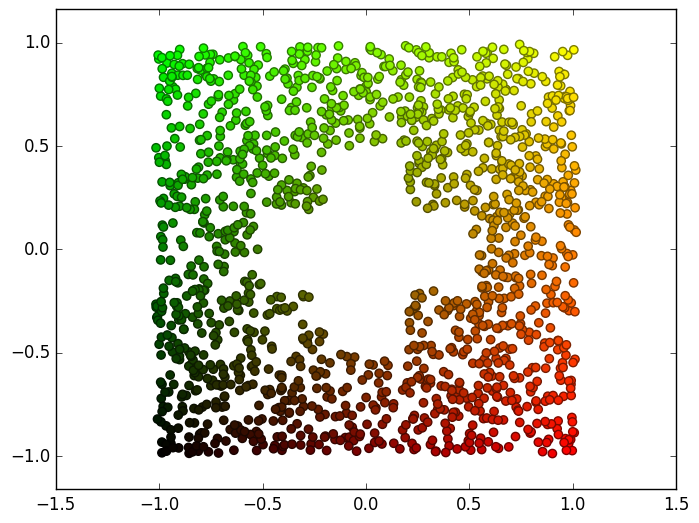}
				\caption{\label{fig:punctured_severed_sphere_intrinsic_isometric_embedding}}
			\end{subfigure}
		\end{centering}
		\begin{centering}
			\begin{subfigure}[b]{0.32\linewidth}
				\includegraphics[width=1\linewidth]{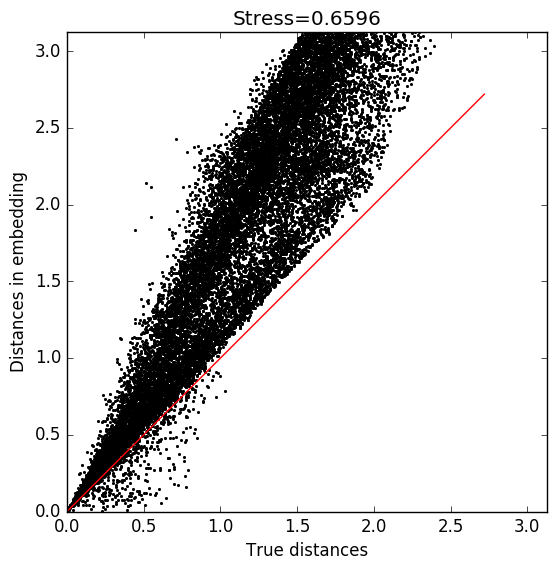}
				\caption{\label{fig:punctured_severed_sphere_standard_isomap_stress}}
			\end{subfigure}
			\hfill
			\begin{subfigure}[b]{0.32\linewidth}
				\includegraphics[width=1\linewidth]{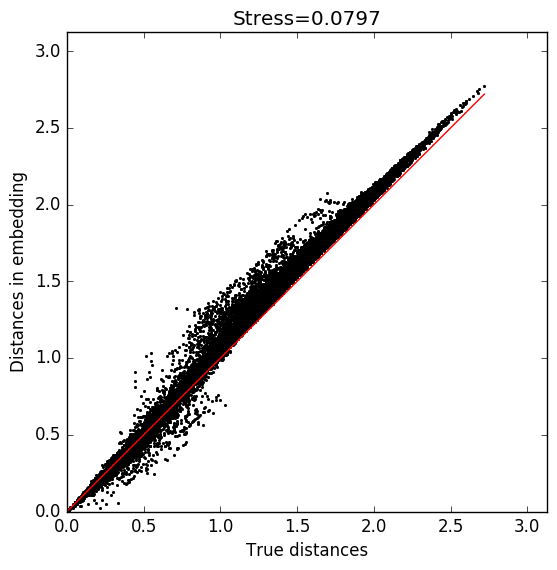}
				\caption{\label{fig:punctured_severed_sphere_intrinsic_isomap_stress}}
			\end{subfigure}
			\hfill
			\begin{subfigure}[b]{0.32\linewidth}
				\includegraphics[width=1\linewidth]{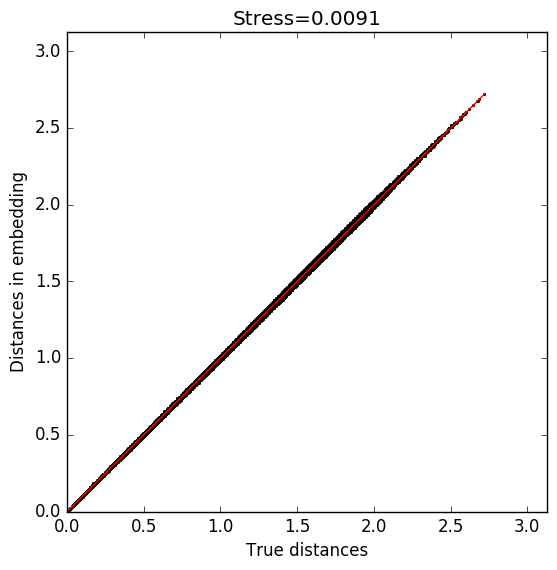}
				\caption{\label{fig:punctured_severed_sphere_intrinsic_isometric_stress}}
			\end{subfigure}
		\end{centering}
		\caption{\label{fig:punctured_severed_sphere} Punctured severed sphere (embedding). \protect\subref{fig:punctured_severed_sphere_intrinsic} Intrinsic space. 
		\protect\subref{fig:punctured_severed_sphere_observed} Observed space. 
		\protect\subref{fig:punctured_severed_sphere_intrinsic_metric} Intrinsic metric. 
		\protect\subref{fig:punctured_severed_sphere_intrinsic_dist_est} Intrinsic distance approximation.
		\protect\subref{fig:punctured_severed_sphere_intrinsic_dist_est_knn} Intrinsic distance approximation - $k$-NN only.
		\protect\subref{fig:punctured_severed_sphere_standard_isomap_embedding} Standard Isomap embedding.
		\protect\subref{fig:punctured_severed_sphere_intrinsic_isomap_embedding} Intrinsic Isomap embedding.
		\protect\subref{fig:punctured_severed_sphere_intrinsic_isometric_embedding} Intrinsic Isometric embedding.
		\protect\subref{fig:punctured_severed_sphere_standard_isomap_stress} Standard Isomap stress.
		\protect\subref{fig:punctured_severed_sphere_intrinsic_isomap_stress} Intrinsic Isomap stress.
		\protect\subref{fig:punctured_severed_sphere_intrinsic_isometric_stress} Intrinsic Isometric stress.
		}
	\end{figure}

	The results of applying the proposed intrinsic isometric embedding algorithm to this data set are displayed in \cref{fig:punctured_severed_sphere}. In \cref{fig:punctured_severed_sphere_intrinsic} we present the intrinsic latent manifold in the intrinsic space and in \cref{fig:punctured_severed_sphere_observed} we present the observed manifold embedded in the observation space. In \cref{fig:punctured_severed_sphere_intrinsic_metric} we visualize the intrinsic metric by plotting corresponding ellipses at several sample points. These ellipses represent the images, under the observation function, of circles of equal radius in the intrinsic space according to the estimated metric. This visualizes the amount of local stretch and contraction in each direction that the observed manifold experiences with respect to the latent intrinsic manifold. In \cref{fig:punctured_severed_sphere_intrinsic_dist_est} we scatter plot the ground truth intrinsic Euclidean pairwise distances and the approximated pairwise distances using \cref{eq:int_dist_approx}. 
	In \cref{fig:punctured_severed_sphere_intrinsic_dist_est_knn}, we scatter plot the same distances, but restrict the depicted pairs of points to only include the $k$ nearest neighbors in the observation space, which are the only distances taken into account by our proposed algorithm. For this example, we use $k=30$. Since the points in the scatter plot are concentrated near the identity map (colored in red), we see that the distance approximation is indeed valid for short distances for which the manifold is approximately linearly distorted. 

	In order to compare our proposed algorithm to manifold learning methods which are not intrinsic-isometric, we plot for each method the resulting embedding and a scatter plot which for each pair of points compares the Euclidean distance in the resulting embedding to the true intrinsic Euclidean distance as measured in the intrinsic space. Again a concentration of points along the identity map represents a better embedding.  Above this scatter plot we also note the calculated stress value according to \eqref{eq:stress} which serves as a quantitative measure of the embedding success. These visualizations are calculated and plotted for standard Isomap (\cref{fig:punctured_severed_sphere_standard_isomap_embedding} and \cref{fig:punctured_severed_sphere_standard_isomap_stress}, intrinsic Isomap (\cref{fig:punctured_severed_sphere_standard_isomap_embedding} and \cref{fig:punctured_severed_sphere_intrinsic_isomap_stress}) and for our proposed intrinsic-isometric algorithm (\cref{fig:punctured_severed_sphere_intrinsic_isometric_embedding} and \cref{fig:punctured_severed_sphere_intrinsic_isometric_stress}). Since the embedding cost functions for these methods are invariant to ridge rotations of the points, all of these plots were registered to the ground truth data for visualization purposes.
	
	As can be clearly observed from these results, our method successfully recovers the structure of the data in the intrinsic space and achieves the minimal stress value. Standard Isomap and the intrinsic variant of Isomap fail to do so and construct a distorted embedding, which does not respect the intrinsic geometric structure, resulting in higher stress values. The failure of standard Isomap in recovering the data can be explained by its dependence on the observation function. Intrinsic Isomap fails due to the non-convexity of the intrinsic manifold and the discrepancy it induces between Euclidean and geodesic intrinsic distances.

	Next, we analyze the effect of using an intrinsic metric which is estimated from the observed data, as opposed to using the exact intrinsic metric. To do so we extend the previous example to include estimation of the metric under the setting described in \cref{ssec:Intrinsic-isotropic-GMM}. Specifically, we use either $N_{i}=5$ or $N_{i}=200$ random measurements sampled from an isotropic Gaussian with intrinsic variance $\sigma_{int}^{2}=0.03^{2}$ centered at each of the data points in the intrinsic space. Additionally, observation noise is added with variance $\sigma_{obs}^{2}=0.03^{2}$.
	
\begin{figure}[H]
		\begin{centering}
			\begin{subfigure}[b]{0.32\linewidth}
					\includegraphics[width=1\textwidth]{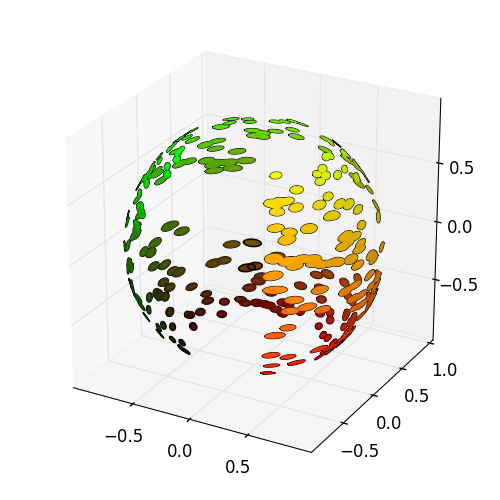}
					\caption{\label{fig:1a}}
			\end{subfigure}\hfill
			\begin{subfigure}[b]{0.32\linewidth}
					\includegraphics[width=1\textwidth]{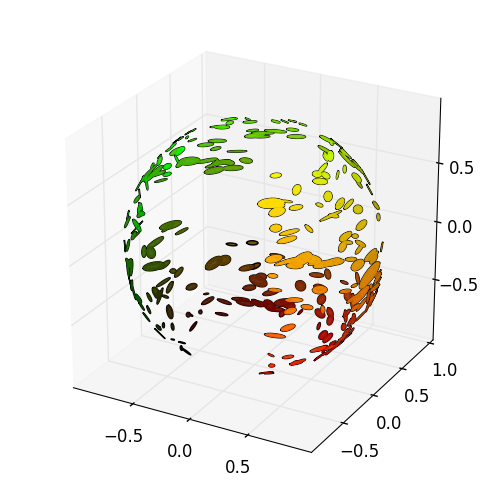}
					\caption{\label{fig:1b}}
			\end{subfigure}\hfill
			\begin{subfigure}[b]{0.32\linewidth}
					\includegraphics[width=1\textwidth]{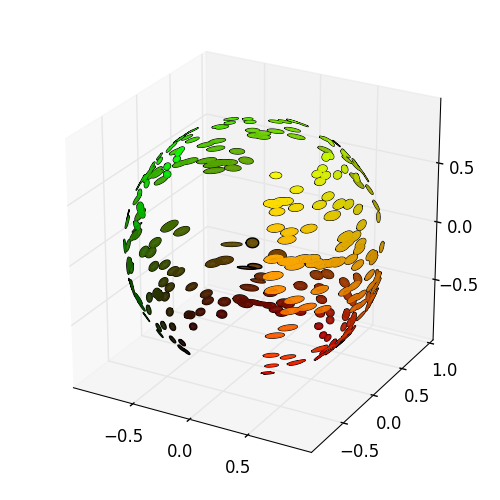}
					\caption{\label{fig:1c}}
			\end{subfigure}
		\end{centering}
		\begin{centering}
			\begin{subfigure}[b]{0.32\linewidth}
					\includegraphics[width=1\textwidth]{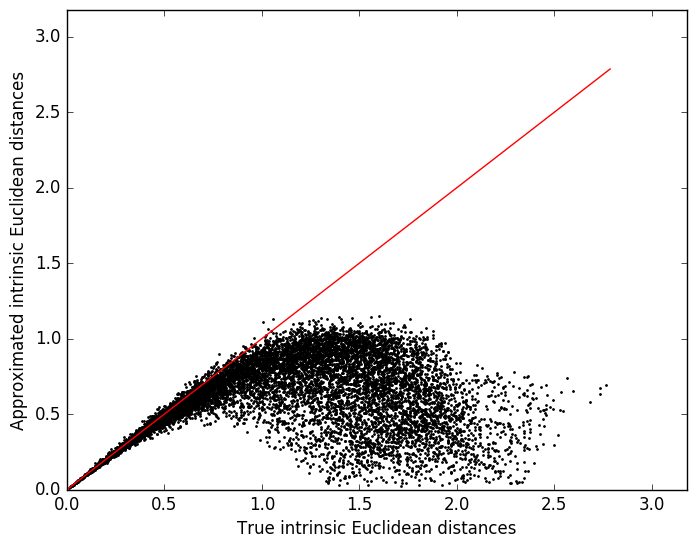}
					\caption{\label{fig:1d}}
			\end{subfigure}\hfill
			\begin{subfigure}[b]{0.32\linewidth}
					\includegraphics[width=1\textwidth]{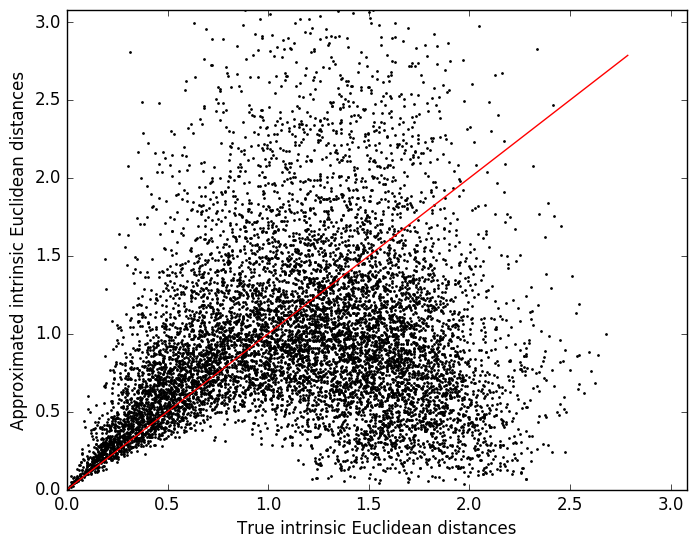}
					\caption{\label{fig:1e}}
			\end{subfigure}\hfill
			\begin{subfigure}[b]{0.32\linewidth}
					\includegraphics[width=1\textwidth]{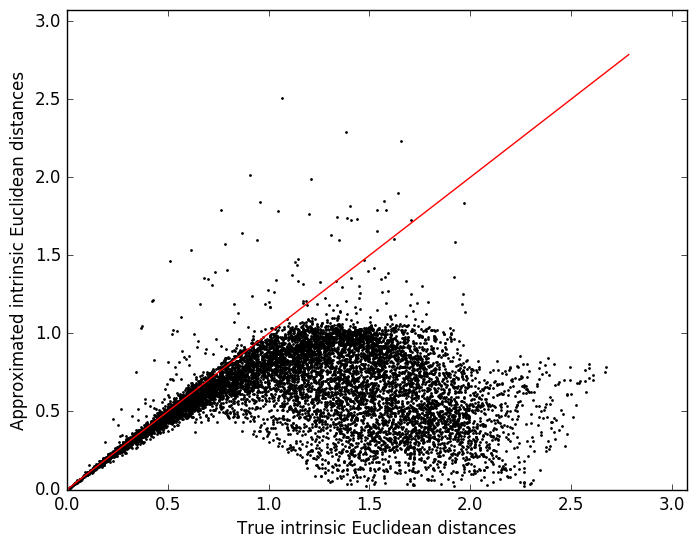}
					\caption{\label{fig:1f}}
			\end{subfigure}
		\end{centering}
		\begin{centering}
			\begin{subfigure}[b]{0.32\linewidth}
					\includegraphics[width=1\textwidth]{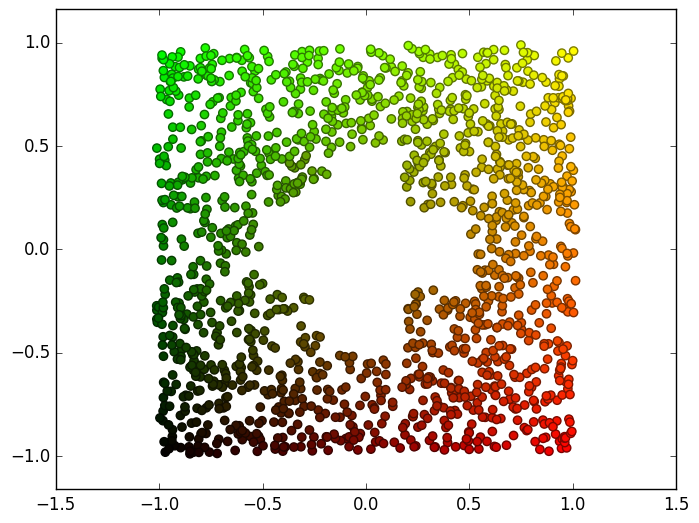}
					\caption{\label{fig:1g}}
			\end{subfigure}\hfill
			\begin{subfigure}[b]{0.32\linewidth}
					\includegraphics[width=1\textwidth]{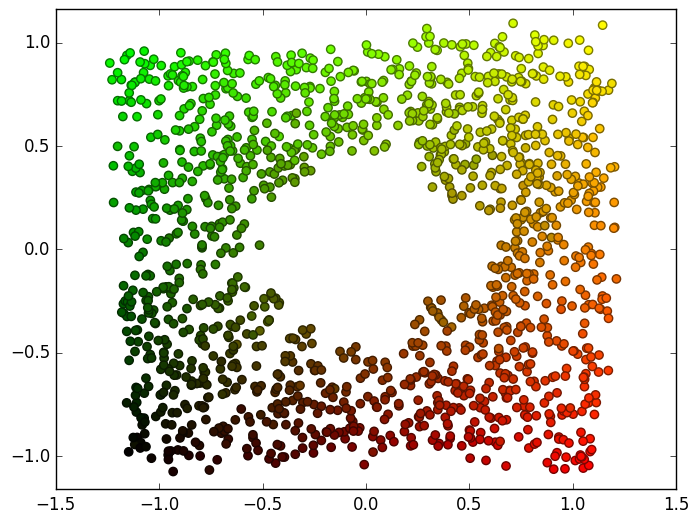}
					\caption{\label{fig:1h}}
			\end{subfigure}\hfill
			\begin{subfigure}[b]{0.32\linewidth}
					\includegraphics[width=1\textwidth]{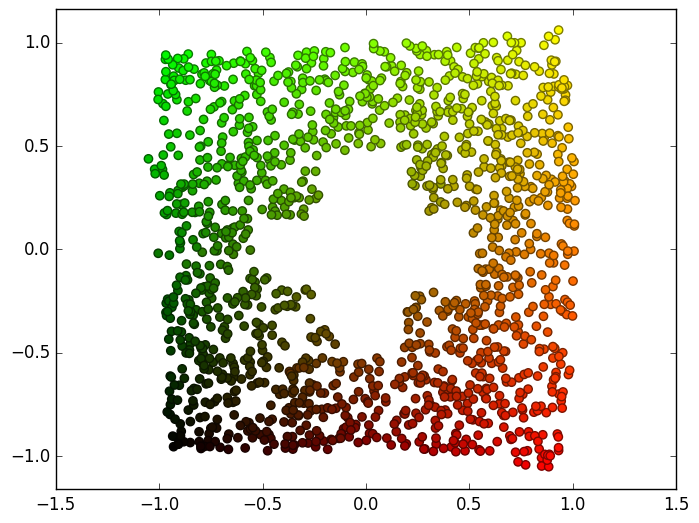}
					\caption{\label{fig:1i}}
			\end{subfigure}
		\end{centering}
		\begin{centering}
			\begin{subfigure}[b]{0.32\linewidth}
					\includegraphics[width=1\textwidth]{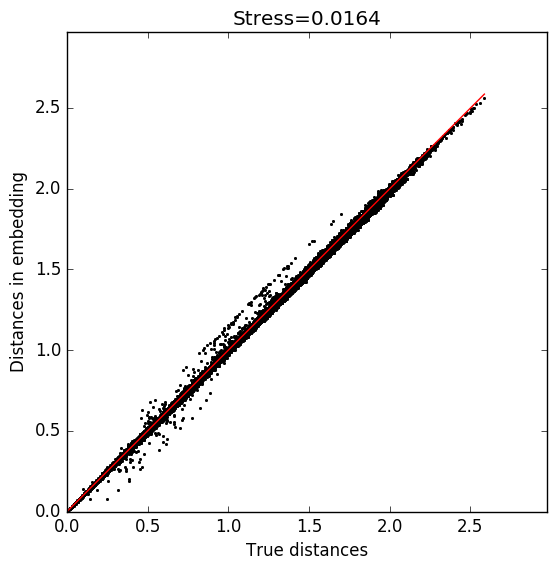}
					\caption{\label{fig:1j}}
			\end{subfigure}\hfill
			\begin{subfigure}[b]{0.32\linewidth}
					\includegraphics[width=1\textwidth]{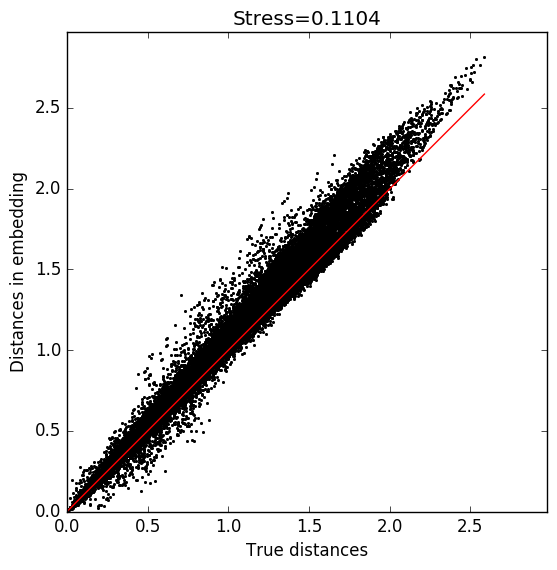}
					\caption{\label{fig:1k}}
			\end{subfigure}\hfill
			\begin{subfigure}[b]{0.32\linewidth}
					\includegraphics[width=1\textwidth]{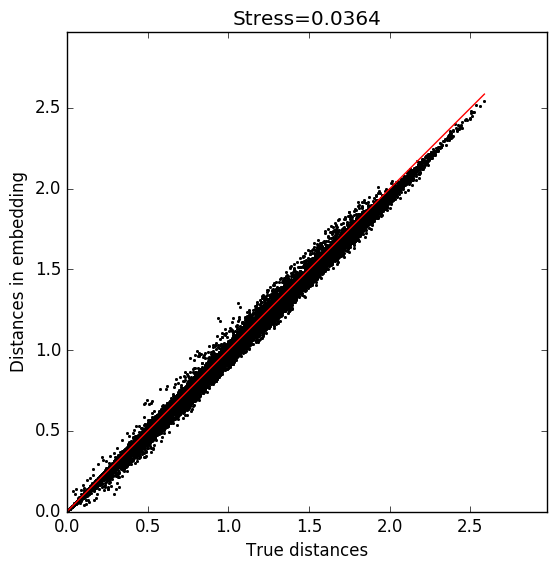}
					\caption{\label{fig:1l}}
			\end{subfigure}
		\end{centering}
	\caption{\label{fig:punctured_severed_sphere_metric} Punctured severed sphere (metric estimation).
		\protect\subref{fig:1a} True intrinsic metric.
		\protect\subref{fig:1b} Locally estimated intrinsic metric.
		\protect\subref{fig:1c} Net learned intrinsic metric.
		\protect\subref{fig:1d} Distance estimation using true metric.
		\protect\subref{fig:1e} Distance estimation using locally estimated metric.
		\protect\subref{fig:1f} Distance estimation using net estimated metric.
		\protect\subref{fig:1g} Embedding using true intrinsic metric.
		\protect\subref{fig:1h} Embedding using locally learned intrinsic metric.
		\protect\subref{fig:1i} Embedding using net learned intrinsic metric.
		\protect\subref{fig:1j} Embedding stress using true intrinsic metric.
		\protect\subref{fig:1k} Embedding stress using locally estimated intrinsic metric.
		\protect\subref{fig:1l} Embedding stress using net learned intrinsic metric}

	\label{fig:metric_punctured_severed_sphere}
	\end{figure}

	In \cref{fig:metric_punctured_severed_sphere} we compare the trivial local estimation (as described in \cref{ssec:Intrinsic-isotropic-GMM}) and the global estimation approach implemented via an \ac{ANN}, as proposed in \cref{sec:intrinsic_metric_estimation}. For each metric estimation method, we plot in \cref{fig:punctured_severed_sphere_metric}, similarity to the previous example: \cref{fig:1a}, \cref{fig:1b} and \cref{fig:1c} a visualization of the metric using ellipses, \cref{fig:1d}, \cref{fig:1e} and \cref{fig:1f} a scatter plot of the true intrinsic distance against the approximated intrinsic distances, \cref{fig:1g}, \cref{fig:1h} and \cref{fig:1i} the resulting low-dimensional embedding, and \cref{fig:1j}, \cref{fig:1k} and \cref{fig:1l} the scatter plot of the Euclidean distances in the embedding compared to the true intrinsic distances, including a calculation of the actual stress value. These are attained for the case of local estimation with dense sampling $N_{i}=200$ (\cref{fig:1a}, \cref{fig:1d}, \cref{fig:1g}, \cref{fig:1j}), for the case of local estimation with sparse sampling $N_{i}=5$ (\cref{fig:1b}, \cref{fig:1e}, \cref{fig:1h}, \cref{fig:1k}) and finally for our proposed \ac{ANN} estimation method with sparse sampling $N_{i}=5$ (\cref{fig:1c}, \cref{fig:1f}, \cref{fig:1i}, \cref{fig:1l}).
	
	
	It can be seen that the local metric estimations in the sparse case are ``noisy'', when compared to those estimated in the dense sampling case, and can sometimes change abruptly between nearby locations on the manifold. This inaccuracy in the estimated local metrics adversely effects the intrinsic distance estimation, and in turn, hinders the recovery of the intrinsic-isometric learned representation. 
	
	It is evident that our neural network based estimation method outperforms the local metric estimation and enables us to learn intrinsic-isometric representation given sparse sampling and in the presence of observation noise. 
	
	
	\subsection{Localization from image data}
	\label{ssec:localization}
	
		In \cref{sec:toy_problem_and_motivation} we provided initial motivation for our work through the simple and intuitive example of localization using observations acquired via an unknown model. We now revisit this example and discuss in detail how the algorithm proposed in this work can be applied to the problem of positioning using image based measurements. Through this problem we examine the advantages of intrinsic geometry preservation and demonstrate its relevance to high-dimensional and complex scenarios.
		
\begin{figure}[h]
			\begin{centering}
				\includegraphics[width=0.7\textwidth]{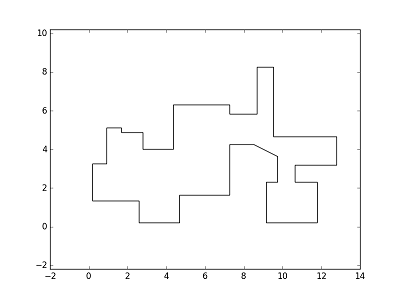}
				\par\end{centering}
			\caption{Outline of the latent manifold which represents a confined indoor environment \label{fig:Indoor-environment-shape}}
		\end{figure}

		The experimental setting is as follows: an agent is located within a compact, path-connected subset $\mathcal{X}$ of $\mathbb{R}^{2}$ which represents a closed indoor environment. The shape of $\mathcal{X}$ used in this experiment is depicted in \cref{fig:Indoor-environment-shape}. At each sample point $\mathbf{x}\in\text{\ensuremath{\mathcal{X}}}$, an image is acquired by the agent. These images serve as an indirect observation or measurement of the location of the agent. Such observations are made in a sufficient number of different locations so that $\mathcal{X}$ is completely covered. The dimension of the intrinsic manifold for this problem is $n=2$, corresponding to the two dimensional physical space and the dimension of the observation space corresponds to the dimensionality of the generated images and is in general high-dimensional. 
	
\begin{figure}[h]
			\begin{centering}
				\begin{subfigure}[b]{0.45\columnwidth}%
					\includegraphics[width=1\textwidth]{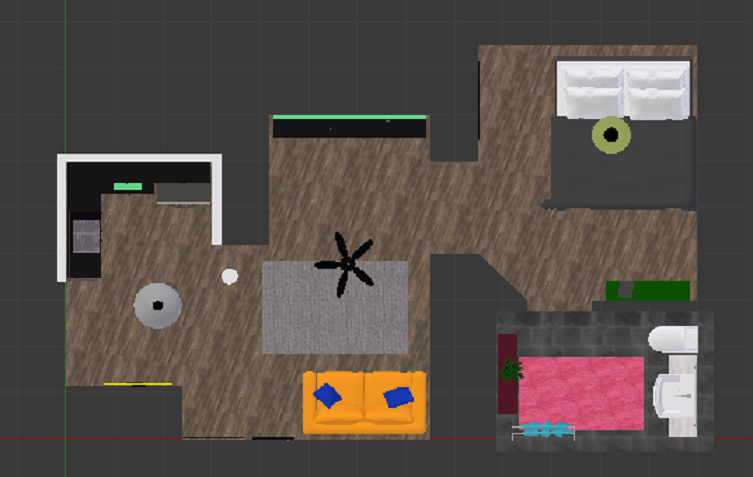}
				\end{subfigure} \hfill
				\begin{subfigure}[b]{0.45\columnwidth}%
					\includegraphics[width=1\textwidth]{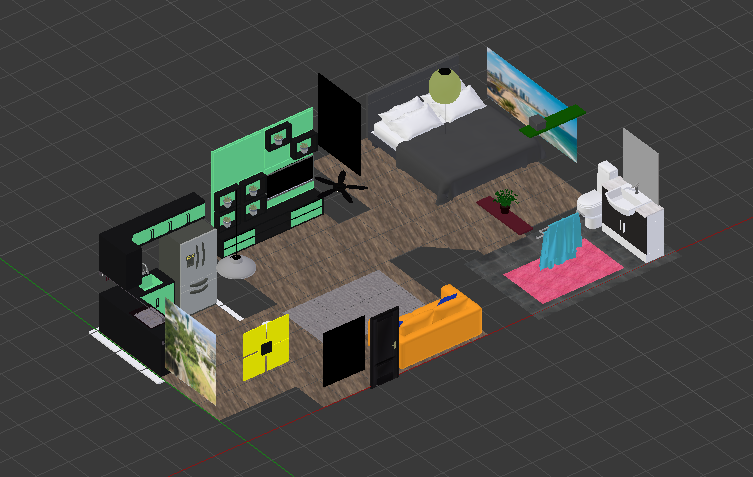}
				\end{subfigure}
			\end{centering}
			\caption{3-dimensional model in Blender\label{fig:3D-model-in}}
		\end{figure}

\begin{figure}[h]
		\begin{centering}
			\begin{subfigure}[t]{0.47\columnwidth}
				\includegraphics[width=1\textwidth]{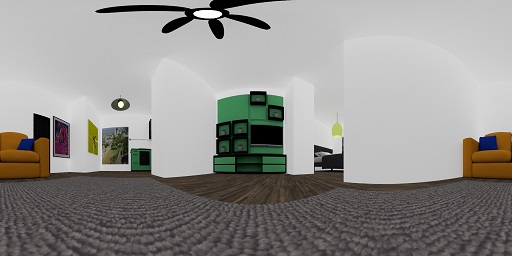}
			\end{subfigure}\hfill
			\begin{subfigure}[t]{0.47\columnwidth}
				\includegraphics[width=1\textwidth]{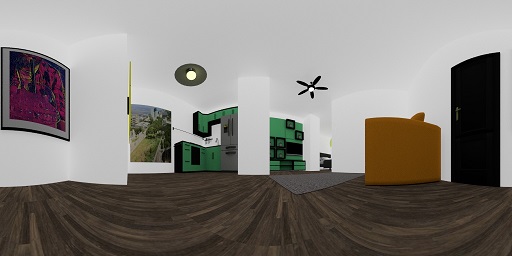}
			\end{subfigure}
		\end{centering}
		
		\begin{centering}
			\begin{subfigure}[t]{0.47\columnwidth}
				\includegraphics[width=1\textwidth]{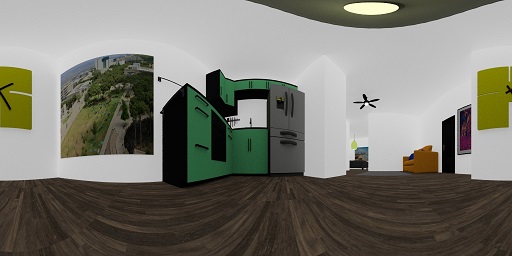}
			\end{subfigure}\hfill
			\begin{subfigure}[t]{0.47\columnwidth}
				\includegraphics[width=1\textwidth]{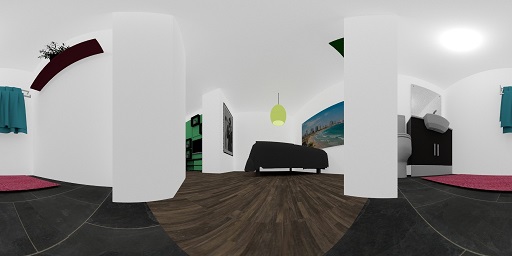}
			\end{subfigure}
		\end{centering}
		\caption{Samples of generated panoramic color images}
		\label{fig:Room-from-Agent's}
		\end{figure}

		To simulate this setting, we used ``Blender''\footnotemark, a professional, freely available and open-source modeling software. Using ``Blender'', we constructed a 3-dimensional model mimicking the interior of an apartment. The created model is shown in \cref{fig:3D-model-in}. The region of the model in which there are no objects and in which the simulated agent is allowed to move, corresponds to the shape of $\mathcal{X}$ presented in \cref{fig:Indoor-environment-shape}. ``Blender'' allows us to render an image of the 3D model as seen via a virtual camera. Using this ability, we produce a set of 360 degree panoramic color images of size $128\times256$ pixels, taken from the point of view of the agent as seen in \cref{fig:Room-from-Agent's}. 
		
		In order for the observation to depend solely on the location of the agent, as is assumed by our proposed algorithm, we were required to omit the affect of the orientation of the agent. This was implemented in the frequency domain by applying a Fourier transform to each frame and then estimating and removing the linear phase in the horizontal direction. Since cyclic rotations of images are equivalent to an addition of linear phase in the Fourier domain, this makes the observations invariant to the agent orientation. 
		
		In order to reduce the initial dimensionality of the data, \ac{PCA} was applied and only the first 100 principal components were taken. This number of principal components was chosen, since empirically 100 components embodied most of the power of the data.
		
		\footnotetext{\url{www.blender.org}}

\begin{figure}[h]
			\begin{centering}
				\begin{subfigure}[t]{0.47\columnwidth}
					\includegraphics[width=1\textwidth]{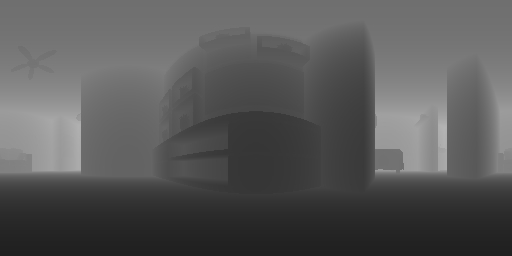}
				\end{subfigure}\hfill
				\begin{subfigure}[t]{0.47\columnwidth}
					\includegraphics[width=1\textwidth]{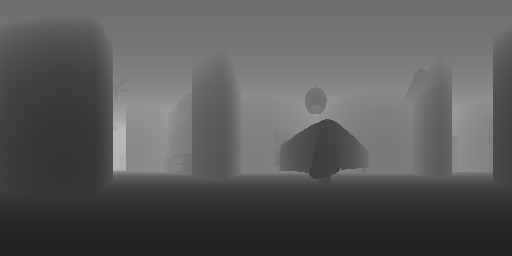}
				\end{subfigure}
			\end{centering}
			
			\begin{centering}
				\begin{subfigure}[t]{0.47\columnwidth}
					\includegraphics[width=1\textwidth]{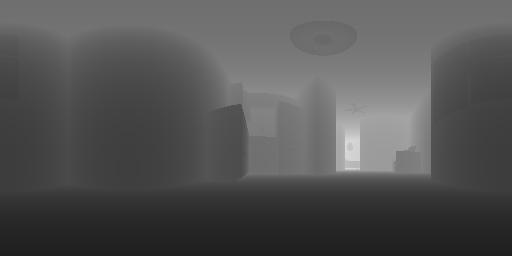}
				\end{subfigure}\hfill
				\begin{subfigure}[t]{0.47\columnwidth}
					\includegraphics[width=1\textwidth]{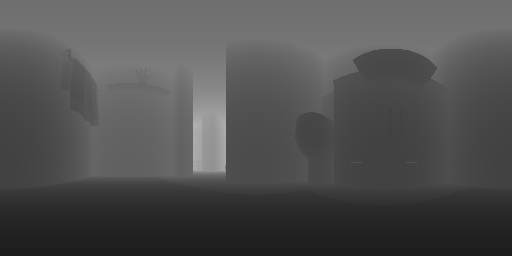}
				\end{subfigure}
			\end{centering}
			\caption{Samples of generated panoramic depth maps \label{fig:Panoramic-depth-images}}
		\end{figure}

		To stress the fact that our algorithm is invariant with respect to the sensor modality, we also used an additional different image modality. The second modality, which is also simulated using ``Blender'', is a gray-scale depth map. For this modality, the gray level at each pixel represents the distance from the camera to the nearest object in the direction of that pixel. Several examples of such images are shown in \cref{fig:Panoramic-depth-images}. This observation modality was also preprocessed similarly to the color images in order to impose invariance to cyclic rotations in the horizontal axis and to preform initial dimensionality reduction.
	
		Images are used as inputs to our algorithm since this represents a possible realistic setting and since it is a non-liner sensor modality which is easy to simulate using 3-dimensional modeling software. We wish to emphasis however, that after the per-processing stage in which these images are made invariant to cyclic rotations, the input is no longer treated as an image, and our proposed algorithm uses no additional image or computer vision related computation on the input. This invariance of the algorithm to the input type enables us, for example, to apply a dimensionality reduction using \ac{PCA} as a preprocessing, despite the fact that it distorts the image structure.
	
		As discussed in \cref{sec:intrinsic_metric_estimation}, in order to uncover the intrinsic metric of the manifold from the observed data, we require the intrinsic data sampling to adhere to some known structure which allows for its estimation from observed data. For this experiment, a different model then the one suggested in \cref{ssec:Intrinsic-isotropic-GMM} however the same neural network parametrization and regularization was used to achieve robustness of the metric estimation.

		For this experiment, the intrinsic sampling is assumed to be acquired by the use of a rigid sensor array. Consider the values of an observation function $f:\mathbb{R}^{n}\to\mathbb{R}^{m}$ observed at $\mathbf{x}+\hat{\mathbf{u}}$ for a point $\mathbf{x}\in\text{\ensuremath{\mathcal{X}}}$ and a small displacement $\hat{\mathbf{u}}$. According to the Taylor series approximation:
		\[
		f\left(\mathbf{x}+\hat{\mathbf{u}}\right)=f\left(\mathbf{x}\right)+\frac{df}{dx}\left(\mathbf{x}\right)\hat{\mathbf{u}}+\mathcal{O}\left(\left\Vert \hat{\mathbf{u}}\right\Vert^2 \right)
		\]
		If the difference between the two measurements $\left\Vert \hat{\mathbf{u}}\right\Vert^2 $ is small with respect to the higher derivatives of the function $f$ we get that:
		\[
		f\left(\mathbf{x}+\hat{\mathbf{u}}\right)-f\left(\mathbf{x}\right)\approx\frac{df}{dx}\left(\mathbf{x}\right)\hat{\mathbf{u}}
		\]
		If one observes the function at $\mathbf{x}$ and at $k$ displacements: $\mathbf{x}+\hat{\mathbf{u}}_{1},\ldots,\mathbf{x}+\hat{\mathbf{u}}_{k}$
		we get a set of equations :
		
		\begin{align*}
			f\left(\mathbf{x}+\hat{\mathbf{u}}_{1}\right)-f\left(\mathbf{x}\right)= & \frac{df}{dx}\left(\mathbf{x}\right)\hat{\mathbf{u}}_{1}\\
			\vdots\\
			f\left(\mathbf{x}+\hat{\mathbf{u}}_{\text{k}}\right)-f\left(\mathbf{x}\right)= & \frac{df}{dx}\left(\mathbf{x}\right)\hat{\mathbf{u}}_{k}
		\end{align*}
		Recasting in matrix form gives:
		\[
		\mathbf{D}=\frac{df}{dx}\left(\mathbf{x}\right)\hat{\mathbf{U}}
		\]
		where
		\begin{align*}
			\hat{\mathbf{U}}= & \left[\hat{\mathbf{u}}_{1}, \cdots, \hat{\mathbf{u}}_{k}\right]\\
			\mathbf{D}= & \left[f\left(\mathbf{x}+\hat{\mathbf{u}}_{1}\right)-f\left(\mathbf{x}\right), \cdots , f\left(\mathbf{x}+\hat{\mathbf{u}}_{k}\right)-f\left(\mathbf{x}\right)\right]
		\end{align*}
		
\begin{figure}[h]
		\begin{centering}
			\includegraphics[width=1\textwidth]{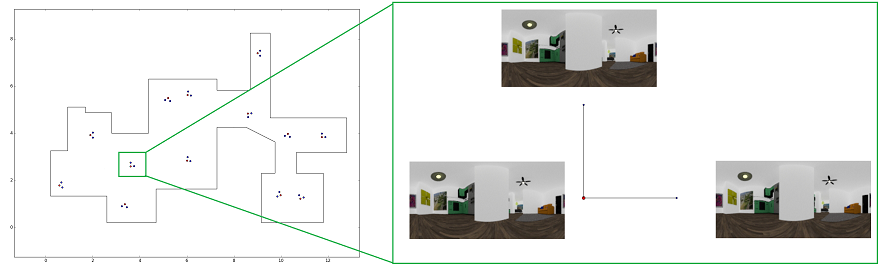}
		\end{centering}
		\caption{Observation points using a rigid sensor array for a subset of 13 sample points. On the right, one can see the effect of slight variations in agents position on the viewed panoramic images \label{fig:sensor_array_localization}}
	\end{figure}

		If the vectors $\hat{\mathbf{u}}_{1},\hat{\mathbf{u}}_{2},\dots,\hat{\mathbf{u}}_{k}$ span the $n$-dimensional intrinsic space (i.e. the matrix $\hat{\mathbf{U}}$ has a full row rank), we can invert this relationship using the pseudo-inverse of $\hat{\mathbf{U}}$ in order to recover the Jacobian of the observation function by:
		\[
		\mathbf{D}\hat{\mathbf{U}}^{T}\left[\hat{\mathbf{U}}\hat{\mathbf{U}}^{T}\right]^{-1}=\frac{df}{dx}\left(\mathbf{x}\right)
		\]
		This suggests a method for estimating the Jacobian of an unknown observation function, using an array of measurements with a known structure. The metric $\mathbf{M}\left(\mathbf{y}\right)$ can then be estimated by:
		\begin{equation}
		\mathbf{M}\left(\mathbf{y}\right)=\frac{df}{dx}\left(\mathbf{x}\right)\frac{df}{dx}\left(\mathbf{x}\right)^{T}\approx\mathbf{D}\hat{\mathbf{U}}^{T}\left[\hat{\mathbf{U}}\hat{\mathbf{U}}^{T}\right]^{-1}\left[\hat{\mathbf{U}}\hat{\mathbf{U}}^{T}\right]^{-T}\hat{\mathbf{U}}\mathbf{D}\label{eq:array-est}
		\end{equation}
		We notice that if this array is rotated around the point $\mathbf{x}$ with a rotation matrix $\mathbf{R}$ we get a rotation of the estimated Jacobian:
		\begin{align*}
			\mathbf{D}\left[\mathbf{R}\hat{\mathbf{U}}\right]^{T}\left[\mathbf{R}\hat{\mathbf{U}}\left[\mathbf{R}\hat{\mathbf{U}}\right]^{T}\right]^{-1} & =\mathbf{D}\hat{\mathbf{U}}^{T}\mathbf{R}^{T}\left[\mathbf{R}\hat{\mathbf{U}}\hat{\mathbf{U}}^{T}\mathbf{R}^{T}\right]^{-1}\\
			& =\mathbf{D}\hat{\mathbf{U}}^{T}\mathbf{R}^{T}\mathbf{R}^{-T}\left[\hat{\mathbf{U}}\hat{\mathbf{U}}^{T}\right]^{-1}\mathbf{R}^{-1}\\
			& =\mathbf{D}\hat{\mathbf{U}}^{T}\left[\hat{\mathbf{U}}\hat{\mathbf{U}}^{T}\right]^{-1}\mathbf{R}^{-1}\\
			& =\frac{df}{dx}\left(\mathbf{x}\right)\mathbf{R}^{-1}
		\end{align*}
		However the estimation of $\mathbf{M}\left(\mathbf{y}\right)$ in \cref{eq:array-est} is unaffected by this rotation, since:
		\begin{equation}
		\frac{df}{dx}\left(\mathbf{x}\right)\mathbf{R}^{-1}\left[\frac{df}{dx}\left(\mathbf{x}\right)\mathbf{R}^{-1}\right]^{T}=\frac{df}{dx}\left(\mathbf{x}\right)\mathbf{R}^{-1}\mathbf{R}^{-T}\cdot\left[\frac{df}{dx}\left(\mathbf{x}\right)\right]=\frac{df}{dx}\left(\mathbf{x}\right)\frac{df}{dx}\left(\mathbf{x}\right)^{T}
		\end{equation}
		This shows that the estimation proposed in \cref{eq:array-est} is invariant to a rigid rotation of the measurement directions $\hat{\mathbf{u}}_{1},\ldots,\hat{\mathbf{u}}_{k}$.
		The invariance to rotations makes this measurement setting practical for intrinsic metric estimation, since the sensor array at different points can be rotated and the respective observations do need to be aligned. In our experiment, the sensor array has an `L' shape and consists of $3$ sensors with $15cm$ distance between measurements as illustrated in \cref{fig:sensor_array_localization}.
	
		Slight variations in the location of the agent cause slight variations in the point of view of the camera and therefore in the rendered image, as seen in \cref{fig:sensor_array_localization}. These slight observed variations, combined with our assumption about the intrinsic structure of the data allow us to infer the local intrinsic metric as described in \cref{sssec:metric_estimation_rigid_array}.
	
%
	
			
\begin{figure}[h]
		\begin{centering}
			\begin{subfigure}[t]{0.47\columnwidth}%
				\includegraphics[width=1\textwidth]{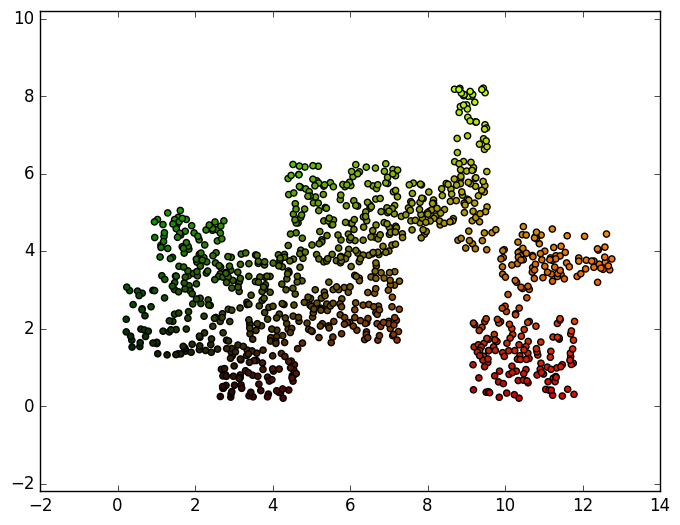}
				\caption{\label{fig:3a}}
			\end{subfigure}\hfill
			\begin{subfigure}[t]{0.47\columnwidth}%
				\includegraphics[width=1\textwidth]{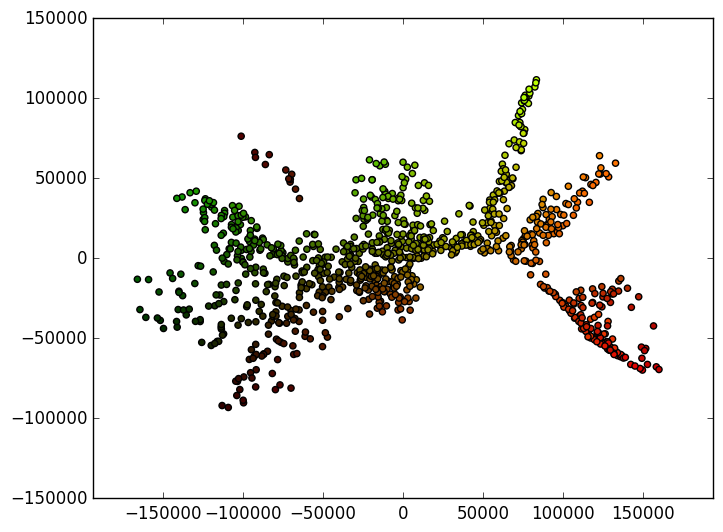}
				\caption{\label{fig:3b}}
			\end{subfigure}
		\end{centering}
		\begin{centering}
			\begin{subfigure}[t]{0.47\columnwidth}%
				\includegraphics[width=1\textwidth]{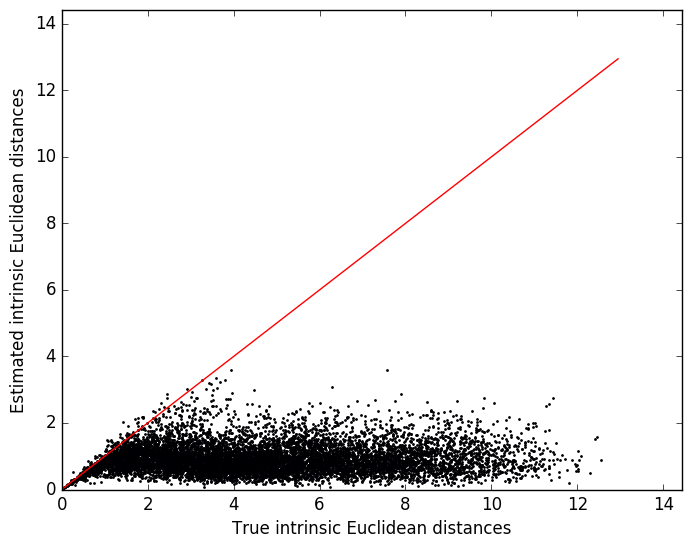}
				\caption{\label{fig:3c}}
			\end{subfigure}\hfill
			\begin{subfigure}[t]{0.47\columnwidth}%
				\includegraphics[width=1\textwidth]{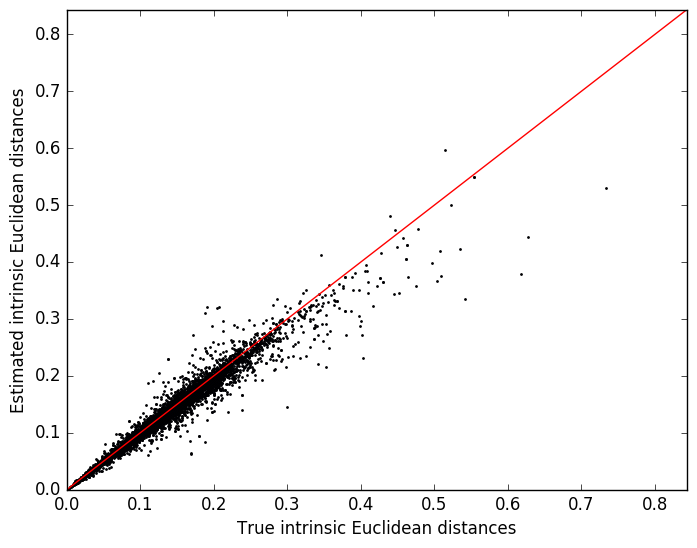}
				\caption{\label{fig:3d}}
			\end{subfigure}
		\end{centering}
		\begin{centering}
			\begin{subfigure}[t]{0.47\columnwidth}%
				\includegraphics[width=1\textwidth]{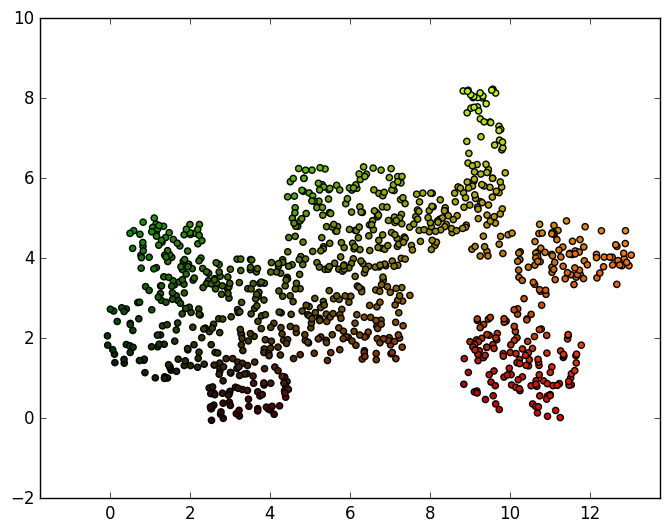}
				\caption{\label{fig:3e} \label{fig:bend1}}
			\end{subfigure}\hfill
			\begin{subfigure}[t]{0.47\columnwidth}%
				\includegraphics[width=1\textwidth]{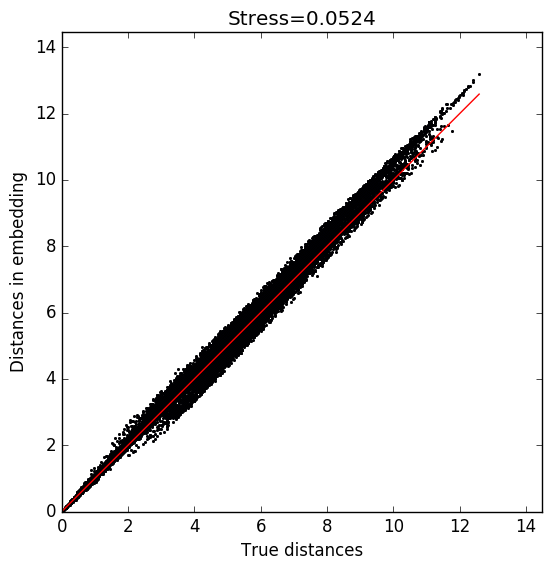}
				\caption{\label{fig:3f}} 
			\end{subfigure}
		\end{centering}
		\caption{Color image observations\label{fig:Color-image-observations}. 
			\protect\subref{fig:3a} Intrinsic space. 
			\protect\subref{fig:3b} Embedding using standard Isomap. 
			\protect\subref{fig:3c} Intrinsic Euclidean distance estimation. 
			\protect\subref{fig:3d} Intrinsic Euclidean distance estimation ($k$-NN). 
			\protect\subref{fig:3e} Intrinsic-isometric embedding. 
			\protect\subref{fig:3f} Euclidean distance discrepancy and stress in resulting embedding}
	\end{figure}

\begin{figure}[h]
		\begin{centering}
			\begin{subfigure}[t]{0.47\columnwidth}
				\includegraphics[width=1\textwidth]{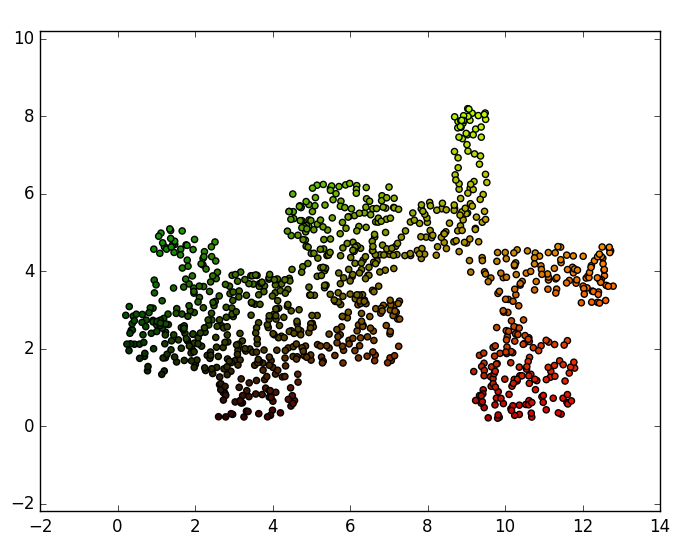}
				\caption{\label{fig:4a}}
			\end{subfigure}\hfill
			\begin{subfigure}[t]{0.47\columnwidth}
				\includegraphics[width=1\textwidth]{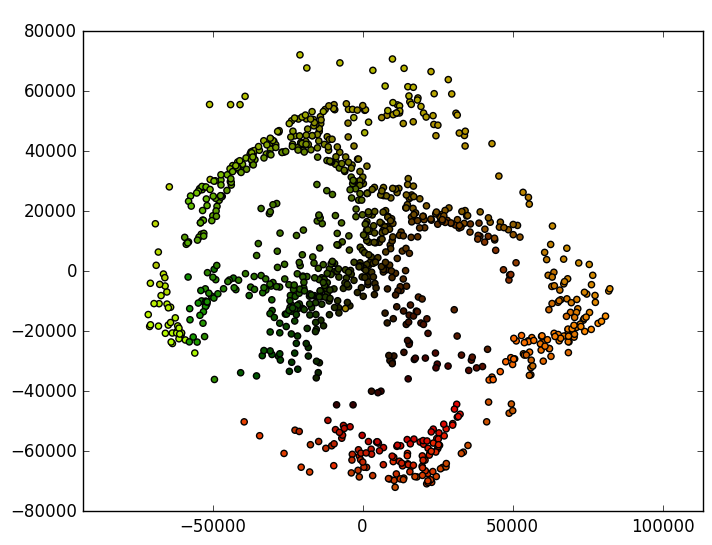}
				\caption{\label{fig:4b}}
			\end{subfigure}
		\end{centering}
		\begin{centering}
			\begin{subfigure}[t]{0.47\columnwidth}
				\includegraphics[width=1\textwidth]{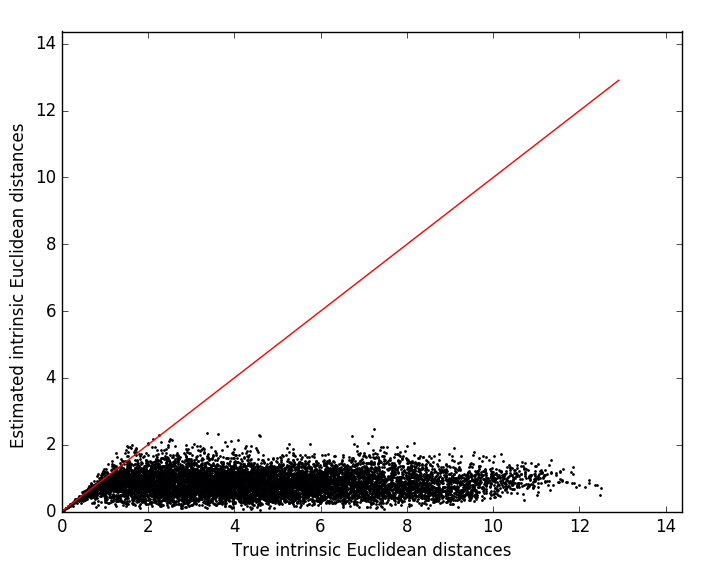}
				\captionsetup{justification=centering}
				\caption{\label{fig:4c}}
			\end{subfigure}\hfill
			\begin{subfigure}[t]{0.47\columnwidth}
				\includegraphics[width=1\textwidth]{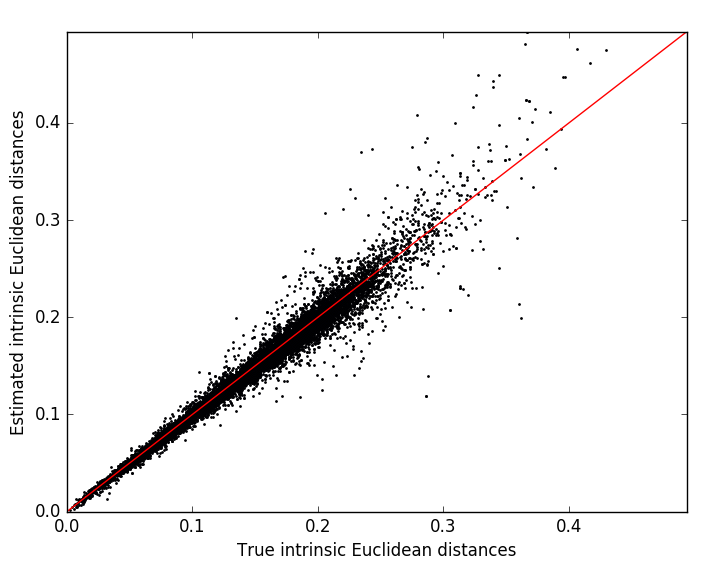}
				\caption{\label{fig:4d}}
			\end{subfigure}
		\end{centering}
		\begin{centering}
			\begin{subfigure}[t]{0.47\columnwidth}%
				\includegraphics[width=1\textwidth]{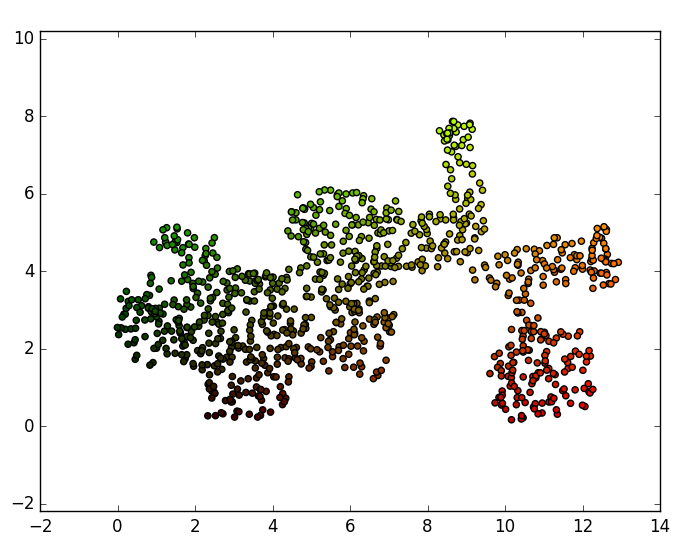}
				\caption{\label{fig:4e} \label{fig:bend2}}
			\end{subfigure}\hfill
			\begin{subfigure}[t]{0.47\columnwidth}
				\includegraphics[width=1\textwidth]{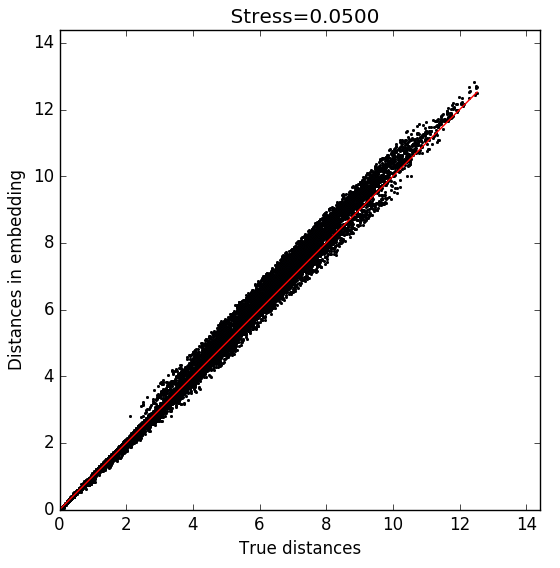}
				\caption{\label{fig:4f}}
			\end{subfigure}
		\end{centering}
		\caption{Depth image observations \label{fig:Depth-image-observations}. 
			\protect\subref{fig:4a} Intrinsic space. 
			\protect\subref{fig:4b} Embedding using standard Isomap. 
			\protect\subref{fig:4c} Intrinsic Euclidean distance estimation. 
			\protect\subref{fig:4d} Intrinsic Euclidean distance estimation ($k$-NN). 
			\protect\subref{fig:4e} Intrinsic-isometric embedding. 
			\protect\subref{fig:4f} Euclidean distance discrepancy and stress in resulting embedding}
	\end{figure}

	For both modalities, $N=1000$ locations were sampled and 3 measurements were made at each such location using the sensor array described above. A 2-dimensional embedding was then constructed based on these measurements and using the proposed algorithm. The constructed embedding was compared to the embedding obtained using the standard Isomap method. Results are presented in \cref{fig:Color-image-observations}, \cref{fig:Depth-image-observations}.
			
	The proposed algorithm accurately retrieves the intrinsic structure for the two observation modalities. This is evident by the recovered intrinsic-isometric embedding structure, which is almost identical to the structure of the points in the latent space. This can be observed for both modalities by comparing \cref{fig:3a} to \cref{fig:3e} and by comparing \cref{fig:4a} to \cref{fig:4e}, as well as from the fact that most pairwise distances in the final embedding closely approximate the true intrinsic Euclidean distances. This also leads to a low stress value for the embedding (as can be observed for both modalities in \cref{fig:3f} and \cref{fig:4f}). The success of the embedding stems from the fact that the estimated intrinsic metric allows for a good estimation of short-range intrinsic distances (as can be observed for both the modalities in \cref{fig:3c}, \cref{fig:3d}, \cref{fig:4c} and \cref{fig:4d}). We notice that since the observation functions used are not locally isometric, standard Isomap fails to retrieve the intrinsic structure of the data or to even provide a 2-dimensional parameterization of the latent space (as can be observed for the two modalities in \cref{fig:3b} \cref{fig:4b}). Isomap does, for the most part, preserve proximity on a local scale (can be observed by points with similar colors being embedded close to each other), yet the global structure is not preserved (as can be observed for the two modalities by comparing \cref{fig:3a} and \cref{fig:3b} and \cref{fig:4a} and \cref{fig:4b}). Since standard Isomap is non-intrinsic, it is effected by the modality of the observation and we receive a different embeddings for different sensor modalities (as can be observed by comparing \cref{fig:3b} and \cref{fig:4b}).
	
	One noticeable weak point of our algorithm, which manifests slightly in these examples, occurs around the coordinate $\left(9,4\right)$ in the true intrinsic space (\cref{fig:3a} and \cref{fig:4a}). This region corresponds to a narrow area in the apartment model in which there are not a lot of sample points.  Typically, errors in distance estimations are attenuated by global self-consistency. However, the distance estimations in this particular region are not sufficiently redundant due to the small number of sample points (and pairwise short distances). This leads to a slight distortion in the embedding of this region, which causes a ``bend'' in the global structure of the embedding (as seen in \cref{fig:bend1} and \cref{fig:bend2}).
	
	These results show that our intrinsic-isometric dimensionality reduction algorithm was successfully applied to the problem of localization and mapping. Basic physical experiments using a randomly walking agent in 2-dimensional space are underway. In these experiments a robotic iRobot Roomba programmable robotic vacuum cleaner was controlled by a Raspberry-Pi mini computer and performed a random walk inside a 2-dimensional region. Signal acquisition was performed either my measuring the \ac{RSS} from WiFi stations or via a wide angle lens which produced 360 degree images. Results from the experiments will be published when concluded.
	
	The advantage of our algorithms is that it exploits a simple prior assumption on the unlabeled measurement acquisition process in order to provide latent system recovery in a manner which is both unsupervised and modality invariant.  These two qualities make it especially suitable for setting where one wants to use an automatic algorithm for localization or/and when the observation model is unknown; which is often the case with indoor localization. However, we wish to remark that we make no claim that this algorithm is superior or even comparable to existing algorithms tailored to specific measurement modalities (if the measurement modality is known a priori) or to machine learning tools trained on labeled data-sets. 

	\subsection{Relation to the sensor network localization problem}
	
	The intrinsicness of our method is achieved by using the push-forward metric for approximation of short-range pairwise distances. Once these distances have been calculated, an embedding of the data into a lower dimensional Euclidean space which respects these distances is constructed. This construction implies that manifold learning problems (with local intrinsic metrics) can be recast as an \ac{SNL} problem \cite{costa2006distributed,yang2009indoor,ash2004sensor,boukerche2007localization,ramadurai2003localization,moses2003self,bal2009localization,biswas2006semidefinite,gepshtein2015sensor,shang2003localization,savvides2001dynamic,savvides2002bits,keller2011diffusion}, in which we are given a number of scattered sensors, each only able to measure distances to a limited number of neighboring sensors, and the task then is to generate a global mapping or embedding of all the sensors in order to accurately localize each sensor. 
			
	Perhaps the leading approach to \ac{SNL} problems is based on relaxation to an \ac{SDP} problem \cite{weinberger2007graph,biswas2006semidefinite,biswas2006semidefinite,biswas2006semidefinite}. This approach suffers from two main shortcomings. Firstly, \ac{SDP} methods scale poorly as the number of points increase, since \ac{SDP} solvers scale cubically with the number of points and constraints. A possible way to reduce the complexity of this approach is to consider representing it in the basis of the leading eigenvectors of the graph Laplacian \cite{weinberger2007graph}. However, this tends to over-smooth the constructed embedding and poorly models the boundaries of the sensors network. A second problem is that the optimal inner-product matrix, found by solving the corresponding \ac{SDP}, is not necessarily of rank corresponding to the intrinsic dimension. To reduce the chance of non-low rank solutions, a maximum-variance regularization term can be added \cite{weinberger2006unsupervised,weinberger2007graph}. This however still does not strictly constrain the rank of the embedding and after the fact dimension truncation of the embedding results in an embedding which violates the distance restrictions.
	
	The \ac{SNL} problem does not have a prevailing solution or method, since each method suffers from some drawbacks. As a result, most methods take a similar approach to the one taken in this paper and regard the constructed embedding only as an initialization point to be further optimized using local descent methods. We note that a similar method  to the one suggested in this paper for construction an embedding once short pairwise distances are given \cite{shang2003localization} appears in. 
	
	The significant difference between \ac{SNL} methods and this work is the broadness of the scope of applicability. For the \ac{SNL} problem the starting point is the availability of true short-range pairwise distances between points in a low-dimensional Euclidean space (calculated by specific hardware in the case of the \ac{SNL} problem), where as in this work the measurement of pairwise distances is not immediate and the main challenge is to first expose the latent low-dimensional Euclidean structure of systems which initially appear to have high-dimensional and complex structure.
	
	\externaldocument{paper.tex}

\section{Conclusions}
\label{sec:conclusions}
		
	In this work, we addressed intrinsic and isometric manifold learning. We first showed that the need for methods which preserve the latent Euclidean geometric structure of data manifolds arises naturally when inherently low-dimensional systems are observed via high-dimensional non-linear measurements. We presented a manifold learning algorithm which uses local properties of the observation function to calculate a local intrinsic metric (the ``push-forward'' metric of the observation function). This metric was used to estimate intrinsic pairwise distances directly. We discussed several settings under which the estimation of the required local observation function properties is possible from the observed data themselves. We recognized that due to their local nature, these metric estimation methods are not sufficiently robust to noise and sparse sampling when the observation function is highly non-linear. To overcome this, we proposed a non-local metric estimation method where an \ac{ANN} was used as a parametric regressor for the intrinsic metric. This computational approach was justified based on a maximum-likelihood estimation under a particular statistical model. We showed that by using smooth non-linearities in the network and additionally restricting the network structure and its weights, we provided sufficient regularization to the estimation, making it robust to noise and sparse sampling. By combining a robust intrinsic metric estimation method and an algorithm which can use these metrics to build an intrinsic and isometric embedding, we devised an algorithm, which can automatically recover the geometric structure of a latent low-dimensional manifold from its observations via an unknown non-linear high-dimensional function. Finally we focused on the example of mapping and positioning an agent using a sensor network of unknown nature and modalities. We showed that our proposed algorithm can recover the structure of the space in which the agent moves and can accurately position the agent within that space. Due to the intrinsic nature of our method, this mapping and positioning do not require prior knowledge of the measurement model. This invariance to the type of measurement used, was shown to be suitable in a setting such as indoor positioning where the exact measurement model is usually unknown.

	
	The manifold learning method proposed in this paper does not require any assumptions be made about the structure of the observed data but rather only on the structure of the data in the intrinsic space. This makes our method invariant to additional deformation and manipulations applied to the observed date. This has the advantage of enabling one to perform pre-processing stages, e.g., for reducing the dimensionality of the data or for removing noise. For example, in \cref{ssec:localization}, when localization and mapping were performed using image data, our algorithm was not applied directly to images, but rather to their low-dimensional projections on the principal components of the data. Importantly, such pre-processing could not be used prior to applications of vision based algorithms, which make assumptions about the measurement model and exploit the structure of image data, since this structure is deformed by the application of \ac{PCA}.

	Another benefit stemming from the invariance to the observation function is the natural ability to facilitate sensor fusion, since in principle, measurements from different sensor modalities can be simply concatenated, creating a higher-dimensional observation function. Although this procedure should work in theory, further research to determine the proper weighting of different measurement modalities is called for. 
		
	
	Supervised machine learning uses labeled data (consisting of pairs of input and desired output values) and attempts to ``learn'' or infer a functional connection between the two. Unfortunately acquisition of labeled data is usually non-trivial, expensive and requires an already existing method to correctly label data. Our method, when viewed end-to-end, is able to label a large set of points by only making an assumption about the intrinsic measurement process. Our proposed approach can therefore be seen as a method for automatic acquisition of labeled data, as it operates in a completely unsupervised manner and produces data labeling as an output. 

	\section*{Acknowledgment}	
	This research was supported by the European Unions Seventh Framework Programme (FP7) under Marie Curie Grant 630657 and by the Technion Hiroshi Fujiwara cyber security research center and the Israel cyber bureau.

	\bibliographystyle{siamplain}
	\bibliography{references}
	
\end{document}